\documentclass[11pt]{article}

\usepackage{appendix}

\usepackage[figuresright]{rotating}

\usepackage{tikz}
\usepackage{pgfplots}
\usepgfplotslibrary{groupplots} 
\usetikzlibrary{pgfplots.groupplots}
\usetikzlibrary{matrix,arrows,decorations.pathmorphing}
\usepackage{pgfplots,pgfplotstable}
\usetikzlibrary{matrix,arrows,decorations.pathmorphing}

\usepackage{enumitem}

\usepackage{amsmath,amssymb,ifthen,color,framed,bm}

\usepackage{graphicx}
\usepackage{caption}
\usepackage[labelformat=simple,subrefformat=simple]{subcaption}

\usepackage{graphicx}
\usepackage{dcolumn}
\usepackage{bm}
\usepackage{amscd,amsthm}
\usepackage{ifthen}
\usepackage{booktabs}
\usepackage{mathtools}

\usepackage{hyperref}
\hypersetup{
    bookmarks=true,         
    unicode=false,          
    pdftoolbar=true,        
    pdfmenubar=true,        
    pdffitwindow=false,     
    pdfstartview={FitH},    
    pdfsubject={Subject},   
    pdfproducer={Producer}, 
    pdfnewwindow=true,      
    colorlinks=true,       
    linkcolor=red,          
    citecolor=blue,        
    filecolor=magenta,      
    urlcolor=cyan           
}

\newtheorem{lemma}{Lemma}

\newtheorem{theorem}{Theorem}

\newtheorem{algorithm}{Algorithm}

\usepackage[
backend=bibtex,
url=false,isbn=false,doi=false,
hyperref=auto,
style=alphabetic,
natbib=true,
sorting=nty,
firstinits=true,
maxnames=2, maxbibnames=10,
]{biblatex}

\bibliography{./Machine_Learning.bib} 

\definecolor{DarkBlue}{rgb}{0,0,0.7} 
\definecolor{BrickRed}{RGB}{203,65,84}

\newcommand\norm[2][\Tnorm]{\ensuremath{{\left\|#2\right\|}_{#1}}}

\newcommand\Tinnerprod{}
\newcommand{\innerprod}[3][\Tinnerprod]{\ifthenelse{\equal{#1}{}}{\ensuremath{\left<#2,#3\right>}}{\ensuremath{\left<#2,#3\right>_{#1}}}}

\newcommand\Tex{}
\newcommand\PR[2][\Tex]{
\ifthenelse{\equal{#1}{}}{{\mathbb P}\left[#2\right]}{\ensuremath{{\mathbb P}_{#1}\left[ #2\right]}}}
\newcommand\EX[2][\Tex]{
\ifthenelse{\equal{#1}{}}{{\mathbb E}\left[#2\right]}{\ensuremath{{\mathbb E}_{#1}\left[ #2\right]}}}

\newcommand\Lnorm[2]{\norm[#2]{#1}}

\newcommand\defeq{\coloneqq}

\newcommand\vect[1]{\ensuremath{#1}}
\newcommand{\vv}{\vect{v}}
\newcommand{\vw}{\vect{w}}
\newcommand{\vz}{\vect{z}}
\newcommand{\vy}{\vect{y}}
\newcommand{\Pmat}{\ensuremath{M}}
\newcommand{\Qmat}{\ensuremath{M'}} 

\newcommand{\reals}{\mathbb R}

\newcommand{\mc}{\mathcal}

\newcommand{\inv}[1]{  {#1}^{ -1 } } 

\long\def\comment#1{}

\renewcommand{\S}{\mathcal S} 

\newcommand\opt[1]{{#1}^*} 

\newcommand\score{\tau}
\newcommand{\vscore}{\tau} 
\newcommand\setb{k} 
\newcommand\numitems{n}
\newcommand{\Shat}{\ensuremath{\widehat{\S}}}
\newcommand{\setind}{\ensuremath{\ell}} 
\newcommand\numsets{\ensuremath{L}} 

\newcommand\alg{\mc A} 
\newcommand\stoptime{\chi} 
\newcommand\numcmp{Q} 

\newcommand{\loindmone}[1]{{\setb_{#1-1}}}      
\newcommand{\loind}[1]{{\setb_{#1-1}+1}}         	
\newcommand{\upind}[1]{\setb_{#1}}			
\newcommand{\upindpone}[1]{{\setb_{#1}+1}}	

\newcommand{\ARloindmone}[1]{{\widehat{\setb}_{#1-1}}}

\newcommand{\ARupindpone}[1]{{\widehat{\setb}_{#1}+1}}	

\newcommand\gapup[2]{\bar{\Delta}_{#1,#2}}
\newcommand\gaplo[2]{{\underline{\Delta}}_{#1,#2}}
\newcommand{\gaplomtwo}[2]{{\underline{\Delta}}^{-2}_{#1,#2}}
\newcommand{\gapupmtwo}[2]{{\bar{\Delta}}^{-2}_{#1,#2}}

\newcommand\timeind{t}

\newcommand\timeup{\overline{\timeind}} \newcommand\timelo{\underline{\timeind}}

\newcommand\bernrv{X} 
\newcommand\kl{d} 

\newcommand{\clobo}{c_{\mathrm{par}}}

\newcommand\complexityP{F} 

\newcommand{\m}{m} 

\newcommand\upindponeph{\upindpone{\setind}}
\newcommand\loindmonemh{\loindmone{\setind}}
\newcommand\ind{i}
\newcommand{\event}{\ensuremath{\mathcal{E}}}

\newcommand{\scorehat}{\ensuremath{\widehat{\score}}}

\newcommand{\MYCDF}{\ensuremath{\Phi}}
\newcommand{\mypdf}{\ensuremath{\phi}}

\newcommand{\sigmaalgebra}{\mc F}

\newcommand{\comparisonclass}[1][]{
\ifthenelse{\equal{#1}{}}{\mc C}{\mc C_{#1}} }
\newcommand{\parfamily}[1][]{
\ifthenelse{\equal{#1}{}}{\mc C_{\mathrm{PAR}}(\MYCDF)}{
\parfamily \cap \comparisonclass[#1]
}}

\newcommand\parw{w} 

\newcommand{\pmat}{\ensuremath{M}}

\newcommand{\pmatmin}{\pmat_{\mathrm{min}}}
\newcommand{\pmatmax}{\pmat_{\mathrm{max}}}

\newcommand{\pdfmin}{\mypdf_{\mathrm{min}}}
\newcommand{\pdfmax}{\mypdf_{\mathrm{max}}}

\newcommand{\comparisonclasstil}{\comparisonclass} 

\newcommand\constup{c_{\mathrm{up}}}
\newcommand\clow{c_{\mathrm{low}}}

\newcommand{\FAR}{\ensuremath{f_{\tiny{\mbox{AR}}}}}
\newcommand{\FLOW}{\ensuremath{f_0}}

\newcommand{\bernrvover}{\ensuremath{\bar{\bernrv}}}

\newcommand{\setbhat}{\ensuremath{\widehat{\setb}}}

\newcommand{\SPECSCORE}[1]{\ensuremath{\{#1\}}}

\newcommand\mystackrel[2]{\stackrel{\text{#1}}{#2}}
\newcommand{\SHORTSCORE}{\ensuremath{\{ \score_i\}_{i=1}^\numitems }}

\newcommand{\mprob}{\ensuremath{\mathbb{P}}}

\makeatletter
\long\def\@makecaption#1#2{
        \vskip 0.8ex
        \setbox\@tempboxa\hbox{\small {\bf #1:} #2}
        \parindent 1.5em  
        \dimen0=\hsize
        \advance\dimen0 by -3em
        \ifdim \wd\@tempboxa >\dimen0
                \hbox to \hsize{
                        \parindent 0em
                        \hfil 
                        \parbox{\dimen0}{\def\baselinestretch{0.96}\small
                                {\bf #1.} #2
                                } 
                        \hfil}
        \else \hbox to \hsize{\hfil \box\@tempboxa \hfil}
        \fi
        }
\makeatother

\newcommand\figureconfidenceintervals{
\begin{tikzpicture}[>=latex,scale=1]

\def\csep{0.5cm}
\def\cright{2.65cm}
\def\cleft{0.7cm}
\def\lenbar{2pt} 

\begin{scope}[xshift=0.5cm]
\node at (-3.5,4) {\parbox{3cm}{\centering latent scores \\of items in $\S_1$}};
\node at (-3.5,2.3) {\parbox{3cm}{\centering latent scores \\of items in $\S_2$}};
\foreach \y/\ytext in { 4.25cm/1, 3.7cm/2, 2.8cm/3,1.75cm/4 }{
   \node[anchor=east] at (-1.5cm,\y) {$\score_{\ytext}$};
   \node[DarkBlue,circle,fill,inner sep=1.2pt] at (-1.5cm,\y) {};
}

\draw [rounded corners,cyan] (-2.15,4.55)--(-1.25,4.55)--(-1.25,3.4)--(-2.15,3.4)--cycle;

\draw [rounded corners,orange] (-2.15,3.1)--(-1.25,3.1)--(-1.25,1.45)--(-2.15,1.45)--cycle;
\end{scope}

\node at (1.75cm,6.2cm) {$t=5$};
\node at (6.5cm,6.2cm) {$t=10$};
\node at (10.25cm,6.2cm) {$t=15$};

\draw[gray, dashed] (2.6cm,4cm) to [out=0,in=180] (4.4cm,4.2cm);
\draw[gray, dashed] (2.6cm,3.2cm) to [out=0,in=180] (4.4cm,3.5cm);
\draw[gray, dashed] (2.6cm,2.8cm) to [out=0,in=180] (4.4cm,2.4cm);
\draw[gray, dashed] (2.6cm,2cm) to [out=0,in=180] (4.4cm,1.85cm);

\begin{scope}
\def\lenconf{1.75cm} 
\foreach \x/\y/\ytext in { 1cm/4cm/1, 1cm + \csep/3.2cm/2, 1cm + 2*\csep/2.8cm/3, 1cm+3*\csep/2cm/4 }{
   \draw[gray] (\cleft,\y) -- (\cright,\y); 
   \node[anchor=east] at (\cleft,\y) {$\scorehat_{(\ytext)}$};
   \node[DarkBlue,circle,fill,inner sep=1.2pt] at (\x,\y) {}; 
   \draw[DarkBlue] (\x,\y-\lenconf) -- (\x,\y+\lenconf);
   \draw[DarkBlue] (\x-\lenbar,\y-\lenconf) -- (\x+\lenbar,\y-\lenconf);
   \draw[DarkBlue] (\x-\lenbar,\y-\lenconf) -- (\x+\lenbar,\y-\lenconf);
   \draw[DarkBlue] (\x-\lenbar,\y+\lenconf) -- (\x+\lenbar,\y+\lenconf);
}
\end{scope}

\draw[gray, dashed] (7.25cm,3.5cm) to [out=0,in=180] (8.85cm,3.7cm);
\draw[gray, dashed] (7.25cm,2.4cm) to [out=0,in=180] (8.85cm,2.8cm);

\begin{scope}[xshift=4.5cm]
\def\lenconf{1.2cm} 
\foreach \x/\y/\ytext in { 1cm/4.2cm/1, 1cm + \csep/3.5cm/2, 1cm + 2*\csep/2.4cm/3, 1cm+3*\csep/1.85cm/4 }{
   \draw[gray] (\cleft,\y) -- (\cright,\y); 
   \node[anchor=east] at (\cleft,\y) {$\scorehat_{(\ytext)}$};
   \node[DarkBlue,circle,fill,inner sep=1.2pt] at (\x,\y) {}; 
   \draw[DarkBlue] (\x,\y-\lenconf) -- (\x,\y+\lenconf);
   \draw[DarkBlue] (\x-\lenbar,\y-\lenconf) -- (\x+\lenbar,\y-\lenconf);
   \draw[DarkBlue] (\x-\lenbar,\y-\lenconf) -- (\x+\lenbar,\y-\lenconf);
   \draw[DarkBlue] (\x-\lenbar,\y+\lenconf) -- (\x+\lenbar,\y+\lenconf);
}
\end{scope}

\begin{scope}[xshift=9cm]
\def\lenconf{0.6cm} 
\foreach \x/\y/\ytext in { 1cm/3.7cm/1, 1cm + \csep/2.8cm/2 }{
   \draw[gray] (\cleft,\y) -- (1.65cm,\y); 
   \node[anchor=east] at (\cleft,\y) {$\scorehat_{(\ytext)}$};
   \node[DarkBlue,circle,fill,inner sep=1.2pt] at (\x,\y) {}; 
   \draw[DarkBlue] (\x,\y-\lenconf) -- (\x,\y+\lenconf);
   \draw[DarkBlue] (\x-\lenbar,\y-\lenconf) -- (\x+\lenbar,\y-\lenconf);
   \draw[DarkBlue] (\x-\lenbar,\y-\lenconf) -- (\x+\lenbar,\y-\lenconf);
   \draw[DarkBlue] (\x-\lenbar,\y+\lenconf) -- (\x+\lenbar,\y+\lenconf);
}
\end{scope}

\end{tikzpicture}
}

\newcommand\figuregaps{
\begin{tikzpicture}[>=latex,scale=1.4]
\def\recw{0.2}
\draw [rounded corners,orange] (2+\recw, -\recw)--(2+\recw,+\recw)--(1-\recw, +\recw)--(1-\recw,-\recw)--cycle;
\draw [rounded corners,cyan] (3.6+\recw, -\recw)--(3.6+\recw,+\recw)--(2.7-\recw, +\recw)--(2.7-\recw,-\recw)--cycle;
\draw [rounded corners,brown] (6+\recw, -\recw)--(6+\recw,+\recw)--(4.5-\recw, +\recw)--(4.5-\recw,-\recw)--cycle;    
\draw (0.4,0.4) -- (6.7,0.4); 
\foreach \x/\xtext in {1/\score_1, 2/\score_2,
  2.7/\score_3, 3.6/\score_4, 4.5/\score_5, 6/\score_6}{ 
  \draw (\x,0.4cm+3pt) -- (\x,0.3);
  \node at (\x,0.2) [anchor=north] {$\xtext$};
  }
   
\draw [decorate,decoration={brace},xshift=0pt,yshift=1.3cm]
(1,0) -- (2.7,0)node [black,midway,yshift=0.4cm] {
$\gaplo{\textcolor{orange}{1}}{1}$
};

\draw [decorate,decoration={brace},xshift=0pt,yshift=0.7cm]
(2,0) -- (2.7,0)node [black,midway,yshift=0.4cm] {
$\gaplo{\textcolor{orange}{1}}{2},\gapup{\textcolor{cyan}{2}}{3}$
};

\draw [decorate,decoration={brace},xshift=0pt,yshift=1.3cm]
(3.6,0) -- (4.5,0)node [black,midway,yshift=0.4cm] {
$\gaplo{\textcolor{cyan}{2}}{4},\gapup{\textcolor{brown}{3}}{5}$
};

\draw [decorate,decoration={brace},xshift=0pt,yshift=0.7cm]
(3.6,0) -- (6,0)node [black,midway,yshift=0.4cm] {
$\gapup{\textcolor{brown}{3}}{6}$
};
\end{tikzpicture}
}

\begin{document}

\begin{center}

{\bf{\LARGE{
Active Ranking from Pairwise Comparisons and \\ 
when Parametric Assumptions Don't Help
}}}

\vspace*{.2in}

{\large{
\begin{tabular}{cccc}
Reinhard Heckel$^{\star}$ & Nihar B. Shah$^{\star}$ & Kannan
Ramchandran$^{\star}$ & Martin J. Wainwright$^{\dagger,\star}$ \\
\end{tabular}
}}

\vspace*{.2in}

\begin{tabular}{c}
Department of Statistics$^\dagger$, and \\
Department of Electrical Engineering and Computer Sciences$^\star$ \\
UC Berkeley,  Berkeley, CA 94720
\end{tabular}

\vspace*{.2in}

\today

\vspace*{.2in}


\begin{abstract}
We consider sequential or active ranking of a set of $\numitems$ items
based on noisy pairwise comparisons. Items are ranked according to the
probability that a given item beats a randomly chosen item, and
ranking refers to partitioning the items into sets of pre-specified
sizes according to their scores.  This notion of ranking includes as
special cases the identification of the top-$k$ items and the total
ordering of the items.  We first analyze a sequential ranking
algorithm that counts the number of comparisons won, and uses these
counts to decide whether to stop, or to compare another pair of items,
chosen based on confidence intervals specified by the data collected
up to that point.  We prove that this algorithm succeeds in recovering
the ranking using a number of comparisons that is optimal up to
logarithmic factors.  This guarantee does not require any structural
properties of the underlying pairwise probability matrix, unlike a
significant body of past work on pairwise ranking based on parametric
models such as the Thurstone or Bradley-Terry-Luce models.  It has been a long-standing open question as to whether or not imposing these parametric assumptions allows for improved ranking algorithms. For stochastic comparison models, in which the pairwise probabilities are bounded away from zero, our second contribution is to resolve this issue by proving a lower bound for parametric models. This shows, perhaps surprisingly, that these popular parametric modeling choices offer at most logarithmic gains for stochastic comparisons.
\end{abstract}

\end{center}


\section{Introduction}

Given a collection of $\numitems$ items, it is frequently of interest
to estimate a ranking based on noisy comparisons between pairs of
items.  Such rank aggregation problems arise across a wide range of
applications.  Some traditional examples in sports include identifying
the best player in a tournament, selecting the top $k$ teams for
playoffs, and finding the full ranking of players.  More recently, the
internet era has led to a variety of applications involving pairwise
comparison data, including recommender
systems~\cite{piech_tuned_2013,aggarwal_recommender_2016} for rating
movies, books, or other consumer items; peer
grading~\cite{shah2013case} for ranking students in massive open
online courses; and online sequential survey
sampling~\cite{salganik_wiki_2015} for assessing the popularity of
proposals in a population of voters. In many of these and other such
applications, it is possible to make comparisons in an active or
adaptive manner---that is, based on the outcomes of comparisons of
previously chosen pairs.  Motivated by those applications, the focus
of this paper is the problem of obtaining statistically sound rankings
based on a sequence of actively chosen pairwise comparisons.

We consider a collection of $\numitems$ items, and our data consists
of outcomes of comparisons between pairs of items in this collection collected
actively. We assume that the outcomes of comparisons are
stochastic---that is, item $i$ beats item $j$ with an unknown
probability $\pmat_{ij} \in (0,1)$. The outcomes of pairwise
comparisons are furthermore assumed to be statistically mutually
independent. We define the ordering of the items in terms of their
(unknown) scores, where the score $\score_i$ of item $i$ is defined as
the probability that item $i$ beats an item chosen uniformly at random
from all other items:
\begin{align}
\label{EqnDefnScore}
\score_i \defeq \frac{1}{\numitems - 1} \sum_{j\neq i} \pmat_{ij}.
\end{align}
In the context of social choice theory~\cite{de1781memoire}, these
sums are also known as the \emph{Borda scores or counts} of the items.
Apart from their intuitive appeal, the Borda counts are of particular
interest because they provide a natural unification of the assumed
orderings in several popular comparison models.  Specifically, the
parametric Bradley-Terry-Luce
(BTL)~\cite{bradley_rank_1952,luce_individual_1959} and
Thurstone~\cite{thurstone_law_1927} models, as well as the
non-parametric Strong Stochastic Transitivity (SST)
model~\cite{tversky_substitutability_1969}, are all based on an
assumed ordering of the items; in all of these models, this ordering
coincides with that given by the scores $\SHORTSCORE$.  In this
paper, we consider the problem of partitioning the items into sets of
pre-specified sizes according to their respective scores.  This notion
of ranking includes as special cases identification of the top-$k$
items and the total ordering of the items.

We make two primary contributions. We begin by presenting and
analyzing a simple active ranking algorithm for estimating a partial
or total ranking of the items. At each round, this algorithm first
counts the number of comparisons won, then computes confidence bounds
from those counts, which it finally uses to select a subset of pairs
to be compared at the next time step.  We provide performance
guarantees showing that with high probability, the algorithm recovers
the desired partial or total ranking from a certain number of
comparisons, which we refer to as the \emph{sample complexity}. We
show that the sample complexity is a function of the (unknown) scores
$\SHORTSCORE$, and therefore distribution-dependent.  Conversely, we
prove distribution-dependent lower bounds that are matching up to
logarithmic factors, thereby showing that the algorithm is
near-optimal in the number of comparisons.  Our analysis leverages the
fact that ranking in terms of the scores $\SHORTSCORE$ is related to a
particular class of multi-armed bandit
problems~\cite{even-dar_action_2006, bubeck_multiple_2013,
  urvoy_generic_2013}. This connection has been observed in past
work~\cite{yuekarmed2012, jamieson_sparse_2015,urvoy_generic_2013} in
the context of finding the top item.

Our second main contribution relates to the popular parametric
modeling choices made in the literature. On one hand, the algorithmic
analysis of this paper does not impose any assumptions on the pairwise
comparison probabilities.  On the other hand, much past work
(including some of our own) is based on specific parametric
assumptions on the pairwise comparisons; for instance, see the
papers~\cite{szorenyi_online_2015, hunter2004mm,
  negahban_iterative_2012, hajek2014minimax, chen2015spectral,
  soufiani2014computing, shah_estimation_2015, maystre2015robust} as
well as references therein.  Concrete examples of parametric assumptions include 
the Bradley-Terry-Luce (BTL) and Thurstone parametric models.  There
is a long standing debate on whether such parametric assumptions are
reasonable---that is, in which situations they (approximately) hold,
and in which they fail~\cite{ballinger_decisions_1997}.  When such
parametric models are suitable, the natural hope is that their 
structure allows some reduction of the sample complexity.  In fact, for
essentially deterministic comparison models (meaning that pairwise comparison
probabilities may be arbitrarily close to zero or one), there can
indeed be significant gains; see the discussion following
Theorem~\ref{thm:necessityparametric} for further details.  However,
as we show in the paper, if one considers \emph{stochastic comparison
  models} (in which the pairwise probabilities are bounded away from
zero and one), then there is at most a logarithmic gain in the sample
complexity in assuming a parametric comparison model over not making any structural assumption.  This
logarithmic gain needs to be weighed against the potential lack of
robustness incurred by using a parametric model 
(note that parametric modeling assumptions often hold only approximately~\cite{ballinger_decisions_1997}, if at all), 
which can be significant, as shown in our numerical results section.


{\bf Related work:} There is a vast literature on ranking and
estimation from pairwise comparison data.  Most works assume
probabilistic comparison outcomes; we refer to the
paper~\cite{jamieson_active_2011} and references therein for ranking
problems assuming deterministic comparison outcomes.  Several prior
works~\cite{hunter2004mm, negahban_iterative_2012, hajek2014minimax,
  soufiani2014computing, shah_estimation_2015, shah_simple_2015,
  chen2016competitive} consider settings where pairs to be compared
are chosen a priori. In contrast, we consider settings where the pairs
may be chosen in an active manner.  The recent
work~\cite{szorenyi_online_2015} assumes the Bradley-Terry-Luce (BTL)
parametric model, and considers the problem of finding the top item
and the full ranking in an active setup.  In the stochastic regime,
for certain underlying distributions, the corresponding
results~\cite[Theorem~3 and Theorem~4]{szorenyi_online_2015} are close
to what our more general result implies. On the other hand, for
several other problem instances the performance guarantees of
Theorem~3 and Theorem~4 in the work~\cite{szorenyi_online_2015} lead
to a significantly larger sample complexity.  Our work thus offers
better guarantees for the BTL model in the stochastic regime, despite
the additional generality of our setting in that we do not restrict
ourselves to the BTL model.  However outside the stochastic regime,
specifically for models with pairwise comparison probabilities very
close to zero and one, \cite[Theorem~3 and
  Theorem~4]{szorenyi_online_2015} offer gains over the results
afforded by our more general model; we discuss this regime in more
detail later.  The paper~\cite{maystre2015robust} considers the
problem of finding a full ranking of items for a BTL pairwise
comparison model, and provides a performance analysis for a
probabilistic model on the BTL parameter vector. Finally,
\citet{eriksson_learning_2013} considers the problem of finding the
very top items using graph based techniques,
\citet{busa-fekete_top-k_2013} consider the problem of finding the
top-k items, and \citet{ailon_active_2011} considers the problem of
linearly ordering the items so as to disagree in as few pairwise
preference labels as possible. Our work is also related to the
literature on multi-armed bandits, and we revisit these relations
later in the paper.

{\bf Organization:} The remainder of this paper is organized as
follows.  We begin with background and problem formulation in
Section~\ref{sec:problemformulation}. We then present a description
and a sharp analysis of our ranking algorithm in
Section~\ref{sec:algo}. In Section~\ref{sec:parametric}, we show that
parametric assumptions do not reduce the sample complexity in the
stochastic regime.  In Section~\ref{sec:simulations} we study
numerically whether algorithms designed for parametric models can
yield some improvement outside the stochastic regime, and study some
additional aspects of our proposed algorithm.  We provide proofs of
all our results in Section~\ref{sec:proofs}, and conclude with a
discussion in Section~\ref{sec:discussion}.


\section{Problem formulation and background}
\label{sec:problemformulation}

In this section, we formally state the ranking problem considered in
this paper and formalize the notion of an active ranking algorithm. We
also formally introduce the class of parametric models in this
section.


\subsection{Pairwise probabilities, scores, and rankings}

Given a collection of items $[\numitems] \defeq
\{1,\ldots,\numitems\}$, let us denote by $\pmat_{ij} \in (0,1)$ the
(unknown) probability that item $i$ wins a comparison with item $j$.
For all items $i$ and $j$, we require that each comparison results in
a winner (meaning that $\pmat_{ij} + \pmat_{ji} = 1$), and we set
$\pmat_{ii} = 1/2$ for concreteness.  For each item $i \in
[\numitems]$, consider the score~\eqref{EqnDefnScore} given as
$\score_i \defeq \frac{1}{\numitems-1} \sum_{j \in [\numitems]
  \backslash \{i\}} \pmat_{ij}$. Note that the (unknown) score
$\score_i \in (0,1)$ corresponds to the probability that item $i$ wins
a comparison with an item $j$ chosen uniformly at random from
$[\numitems] \setminus \{i\}$.

Assuming that the scores are all distinct, they define a unique
ranking of the $\numitems$ items; this (unknown) ranking corresponds
to the permutation $\pi:[\numitems]\rightarrow [\numitems]$ such that
\begin{align*}
\score_{\pi(1)} > \score_{\pi(2)} > \ldots > \score_{\pi(\numitems)}.
\end{align*}
In words, $\pi(i)$ denotes the $i^{th}$ ranked item according to the
scores. A number of ranking problems can be defined in terms of $\pi$:
at one extreme, finding the best item corresponds to determining the
item $\pi(1)$, whereas at the other extreme, finding a complete
ranking is equivalent to estimating $\pi(j)$ for all $j \in
[\numitems]$.  We introduce a general formalism that allows us to
handle these and many other ranking problems.  In particular, given an
integer $\numsets \geq 2$, we let
$\{\setb_\setind\}_{\setind=1}^\numsets$ be a collection of positive
integers such that \mbox{$1 \leq \setb_1 < \setb_2 < \ldots <
  \setb_{\numsets-1} < \setb_\numsets = \numitems$.}  Any such
collection of positive integers defines a partition of $[\numitems]$
into $\numsets$ disjoint sets of the form
\begin{align}
\S_1 \defeq \{\pi(1), \ldots, \pi(\setb_1)\},\;\; \S_2 \defeq
\{\pi(\setb_1+1), \ldots, \pi(\setb_2) \},\ldots,\;\; \S_{\numsets}
\defeq \{ \pi(\setb_{\numsets-1} +1), \ldots, \pi(\numitems)\}.
\label{eq:defsets}
\end{align}
For instance, if we set $\numsets = 2$ and $\setb_1 = k$, then the set
partition $(\S_1, \S_2)$ corresponds to splitting $[\numitems]$ into
the top $k$ items and its complement.  At the other extreme, if we set
\mbox{$\numsets = \numitems$} and \mbox{$(\setb_1, \setb_2, \ldots,
  \setb_\numitems) = (1, 2, \ldots, \numitems)$,} then the partition
$\{\S_\setind\}_{\setind=1}^\numsets$ allows us to recover the full
ranking of the items, as specified by the permutation $\pi$.

For future reference, we define
\begin{align}
\comparisonclass[\pmatmin] & \defeq \left\{\Pmat \in (0,1)^{n\times n}
\mid \pmat_{ij} = 1-\pmat_{ji}, \pmat_{ij} \geq \pmatmin, \text{ and }
\tau_i \neq \tau_j \text{ for all } (i,j) \right\},
\label{eq:defcomparisonclass}  
\end{align}
corresponding to the set of pairwise comparison matrices with pairwise
comparison probabilities lower bounded by $\pmatmin$, and for which a
unique ranking exists.\footnote{We note that our results actually do
  not require the entire underlying ordering of the scores to be
  strict; rather, we require strict inequalities only at the
  boundaries of the sets $\S_1,\ldots,\S_\numsets$.}


\subsection{The active ranking problem}

An active ranking algorithm acts on a pairwise comparison model $\Pmat
\in \comparisonclass[0]$.  Consider any specified values of $\numsets$
and $\{\setb_\setind\}_{\setind=1}^\numsets$ defining a partition of
the form~\eqref{eq:defsets} in terms of their latent
scores~\eqref{EqnDefnScore}. The goal is to obtain a partition of the
items $[\numitems]$ into $\numsets$ disjoints sets of the
form~\eqref{eq:defsets} from active comparisons.  At each time
instant, the algorithm can compare two arbitrary items, and the choice
of which items to compare may be based on the outcomes of previous
comparisons. As a result of comparing two items $i$ and $j$, the algorithm
receives an independent draw of a binary random variable with success
probability $\pmat_{ij}$ in response.  After termination dictated by
an associated stopping rule, the algorithm returns a ranking $\Shat_1,
\ldots, \Shat_\numsets$.

For a given tolerance parameter $\delta \in (0,1)$, we say that a
ranking algorithm $\alg$ is \emph{$\delta$-accurate for a comparison
  matrix $\Pmat$} if the ranking it outputs obeys
\begin{align}
\label{EqnDeltaAccurate}
  \PR[\Pmat]{ \Shat_\setind = \S_\setind, \text{ for all } \setind=1,
    \ldots,\numsets } \geq 1- \delta.
\end{align}
For any set of comparison matrices $\comparisonclasstil$, we say that
the algorithm $\alg$ is \emph{uniformly $\delta$-accurate} over
$\comparisonclasstil$ if it is $\delta$-accurate for each matrix
$\Pmat \in \comparisonclasstil$.  The performance of any algorithm is
measured by means of its \emph{sample complexity}, by which we mean
the number of comparisons required to obtain the desired partition.


\subsection{Active ranking and multi-armed bandits}
\label{sec:relationbandits}

It is worthwhile noting that the ranking problem studied here is
related to multi-armed
bandits~\cite{kaufmann_complexity_2014,bubeck_regret_2012}.  More
precisely, a multi-armed bandit model consists of a collection of
$\numitems$ ``arms'', each associated with an unknown and stochastic
reward function, and the goal is to maximize the reward obtained via a
sequential choice of arms.  In past work, various researchers
(e.g.,~\cite{yue_beat_2011,yuekarmed2012,urvoy_generic_2013,jamieson_sparse_2015})
have drawn links between pairwise comparison ranking and such bandit
problems.  In particular, by definition of the score $\score_i$,
comparing item $i$ to a distinct item chosen from the $\numitems-1$
alternatives can be modeled as drawing a Bernoulli random variable
with mean $\score_i$. Our subsequent analysis in
Section~\ref{sec:algo} relies on this relation.  When cast in the
multi-armed bandit setting, the setting of pairwise comparisons is
often referred to as that of ``dueling bandits''.  Prior works in this
setting~\cite{yue_beat_2011,yuekarmed2012,urvoy_generic_2013,jamieson_sparse_2015}
address the problem of finding the single ``best arm''--- meaning the
item with the highest score---based on noisy comparisons.  By
contrast, this paper treats the more general problem of finding a
partial or total ordering of the items.

Despite these similarities, there is an important distinction between
the two settings.  If we view our problem as a multi-armed bandit
problem with Bernoulli random variables with means
$\{\score_i\}_{i=1}^\numitems$, these means are actually coupled
together, in the sense that information about any particular mean
imposes constraints on all the other means.  In particular, any set of
scores $\{\score_i\}_{i=1}^\numitems$ must be realized by some valid
set of pairwise comparison probabilities $\{\pmat_{ij}\}_{i,j\in
  [\numitems]}$.  Since these pairwise comparison probabilities must
obey the constraint $\pmat_{ij} = 1 - \pmat_{ji}$, the
induced scores must satisfy certain constraints, not all of
which are obvious.  One obvious constraint, which follows immediately
from the definition~\eqref{EqnDefnScore}, is that
$\sum_{i=1}^\numitems \score_{i} = n/2$, another constraint is that 
$\sum_{i=1}^j \score_{\pi(i)} \geq \frac{1}{\numitems - 1} \frac{j (j-1)}{2}$ \cite{landau_dominance_1953,joe_majorization_1988}. 
Those condition, while
necessary, are certainly not sufficient, as can be seen by studying
some simple cases.\footnote{For instance, there is no set of pairwise
  comparison probabilities with scores $[1,1,0,0]$, even though those
  scores satisfy the aforementioned constraints. In order to verify this fact, note that
  $\score_1 = 1$ implies $\pmat_{12}=\pmat_{13}=\pmat_{14} = 1$. Thus,
  we have $\pmat_{21} = 0$, which implies $\score_2 \leq 2/3$ and
  therefore contradicts $\score_2 = 1$.}
  Our algorithm, presented in the next section, does not take the coupling of the scores explicitly into account. 
  Nevertheless, our algorithm is shown to be optimal up to a logarithmic factor in the stochastic regime. 


\subsection{Parametric models}
\label{sec:intro_parametric}

In this section, we introduce a family of parametric models that form
a basis of several prior works~\cite{szorenyi_online_2015,
  hunter2004mm, negahban_iterative_2012, hajek2014minimax,
  shah_estimation_2015}.  To be clear, we make no modeling assumptions
for our algorithm and its analysis in Section~\ref{sec:algo}. Rather,
we focus on these parametric models in Section~\ref{sec:parametric},
where we show that, perhaps surprisingly, outside of the deterministic
regime, none of these parametric assumptions provide more than a
logarithmic gain in sample complexity.

Any member of this family is defined by a strictly increasing and
continuous function $\MYCDF \colon \reals \to [0,1]$ obeying $\MYCDF(
t ) = 1 - \MYCDF(-t)$, for all $t \in \reals$. The function $\MYCDF$
is assumed to be known.  A pairwise comparison matrix in this family
is associated to an unknown vector $\vw \in \reals^\numitems$, where
each entry of $\vw$ represents some quality or strength of the
corresponding item.  The parametric model $\parfamily$ associated with
the function $\MYCDF$ is defined as:
\begin{align}
\label{eq:defn_parmodel}
\parfamily = \{ \pmat_{ij} = \MYCDF(\parw_i - \parw_j) \ \ \mbox{for
  all }i,j \in [\numitems],\ \ \text{ for some } \vw \in
\reals^\numitems \}.
\end{align}

Popular examples of models in this family are the Bradley-Terry-Luce
(BTL) model, obtained by setting $\MYCDF$ equal to the sigmoid
function ($\MYCDF(\timeind) = \frac{1}{1 + e^{-t}}$), and the
Thurstone model, obtained by setting $\MYCDF$ equal to the Gaussian
CDF.  Note that \mbox{$\score_1 > \score_2 > \ldots >
  \score_\numitems$} is equivalent to \mbox{$\parw_1 > \parw_2 >
  \ldots > \parw_\numitems$}, meaning that the ranking induced by the
 scores $\SHORTSCORE$ is equivalent to that induced by $\parw$.

It is worthwhile noting that a common assumption in the setting of
parametric models~\cite{negahban_iterative_2012, shah_estimation_2015,
  chen2015spectral} is that $\Lnorm{\parw}{\infty} \leq B$ for some
finite constant $B$. This boundedness assumption implies that the
pairwise comparison probabilities $ \{\pmat_{ij} \}_{i,j=1}^\numitems$
are all uniformly bounded away from $0$ and $1$, thereby guaranteeing
a stochastic comparison model.


\section{Active ranking from pairwise comparisons}
\label{sec:algo} 

In this section, we present our algorithm for obtaining the desired
partition of the items as described earlier in
Section~\ref{sec:problemformulation}, and a sharp analysis of this
algorithm proving its optimality up to logarithmic factors.

\subsection{Active ranking (AR) algorithm}

Our active ranking algorithm is based two ingredients:
\begin{itemize}
\item Successive estimation of the scores $\SHORTSCORE$, where score
  $\score_i$ is estimated by comparing item $i$ with items chosen
  uniformly at random from $[\numitems] \setminus \{i\}$.
\item Assigning an item $i$ to an estimate $\Shat_\setind$ of the set
  $\S_\setind$ once a certain confidence level of $i$ belonging to
  $\S_\setind$ is attained.
\end{itemize}
This strategy is essentially an adaption of the successive elimination
approach from the bandit literature, proposed in the classical
paper~\cite{paulson_sequential_1964}, and studied in a long line of
subsequent work (see, for
example,~\cite{even-dar_action_2006,bubeck_multiple_2013,urvoy_generic_2013,jamieson_sparse_2015}).

The first input to the algorithm is a collection of positive integers
$\{\setb_\setind\}_{\setind=0}^\numsets$ such that
\begin{align*}
\mbox{$\setb_0 = 0 < \setb_1 < \setb_2 < \ldots < \setb_{\numsets-1} <
  \setb_\numsets = \numitems$,}
\end{align*}
which define a desired ranking.  The second input is a tolerance
parameter $\delta \in (0,1)$, which defines the probability with which
the algorithm is allowed to fail.


\begin{figure}[t]
  \centering \figureconfidenceintervals \captionof{figure}{
\label{fig:confbounds}
Illustration of the AR algorithm applied to the problem of finding the
top $2$ items out of $\numitems=4$ items total, corresponding to $\S_1
= \{1,2\}, \S_2 = \{3,4\}$. The figure depicts the estimates
$\scorehat_i(\timeind)$, along with the corresponding confidence
intervals $[\scorehat_i(\timeind) - 4 \alpha_\timeind,
  \scorehat_i(\timeind) + 4 \alpha_\timeind]$, at different time steps
$t$. At time $t=5$, the algorithm is not confident about the position
of any of the items, and hence it continues to sample further.  At
time $t=10$, the confidence interval of item $(1)$ indicates that
$(1)$ is either the best or the second best item, therefore the AR
algorithm assigns $(1)$ to $\Shat_1$.  Likewise, it assigns item $(4)$
to $\Shat_2$.  At time step $t=15$, the AR algorithm assigns items
$(1)$ and $(2)$ to $\Shat_1$ and $\Shat_2$, respectively, and
terminates. }
\end{figure}


\begin{algorithm}[Active Ranking (AR)]
At time $\timeind=0$, define and initialize the following quantities:
\begin{center}
\begin{tabular}{>{$\bullet$~~~~}lll}
$\S= [\numitems]$ & \qquad\qquad & (set of items not ranked yet);\\
$\Shat_\setind = \emptyset$ \quad \mbox{for all $\setind \in
    [\numsets]$} & & (estimates of the partition);\\
$\setbhat_\setind = \setb_\setind$ \quad \mbox{for all $\setind \in
    \{0,\ldots, \numsets\}$} & & (borders of the sets);\\
$\scorehat_i(0) = 0$ \quad \mbox{for all $i \in [\numitems]$} &&
  (estimates of the scores).
\end{tabular} 
\end{center}

At any time $\timeind\geq 1$:
\begin{enumerate}
\item For every $i \in \S$: Compare item $i$ to an item chosen
  uniformly at random from $[\numitems] \setminus \{i\}$, and set
\begin{align}
\label{eq:scorehatupdate}
\scorehat_i(t) =
\begin{cases}
\frac{t-1}{t} \scorehat_i(t-1) + \frac{1}{t} & \qquad \mbox{if $i$
  wins}\\ \frac{t-1}{t} \scorehat_i(t-1) & \qquad \mbox{otherwise.}
\end{cases}
\end{align}
\item Sort the items in set $\S$ by their current estimates of the
  scores: For any $k \in [|\S|]$, let $(k)$ denote the item with the
  $k$-th largest estimate of the score.
\item With $\alpha_\timeind \defeq \sqrt{\frac{\log(125 \numitems
    \log(1.12 \timeind)/\delta)}{\timeind}}$, do the following for
  every $j \in \S$: \\
If the following pair of conditions~\eqref{eq:bworst}
and~\eqref{eq:bbest} hold simultaneously for some $\ell \in
[\numsets]$,
\begin{subequations}
\begin{align}
& \setbhat_{\setind-1}=0 \quad \text{or} &\scorehat_{j}(\timeind) <
  \scorehat_{( \setbhat_{\setind - 1})}(t) - 4 \alpha_\timeind
\quad &(\text{$j$ likely is one of the lower
  $\numitems-\setb_{\setind-1}-1$ items})
\label{eq:bworst} \\
& \setbhat_{\setind} = |\S| \quad\text{or} &\scorehat_{j}(\timeind) >
\scorehat_{(\setbhat_\setind + 1)}(t) + 4 \alpha_\timeind \quad
&(\text{$j$ likely is one of the top $\setb_{\setind}$ items}),
\label{eq:bbest}
\end{align}
\end{subequations}
then add $j$ to $\Shat_\setind$, remove $j$ from $\S$, and set
$\setbhat_{\setind'} \leftarrow \setbhat_{\setind'} -1$ for all
$\setind' \leq \setind$.
\item 
If $\S = \emptyset$, terminate.
\end{enumerate}
\end{algorithm}

\noindent See Figure~\ref{fig:confbounds} for an illustration of the
progress of this algorithm on a particular instance.


\subsection{Guarantees  and optimality of the AR algorithm}
\label{sec:general_lower}

In this section, we establish guarantees on the number of samples for
the AR algorithm to succeed. As we show below, the sample complexity
is a function of the gaps between the scores, defined as
\begin{align}
\label{EqnGap}
\gapup{\setind}{i} \defeq \score_{\pi(\loindmone{\setind})} -
\score_{i}, \quad \mbox{and} \quad
\gaplo{\setind}{i} \defeq \score_{i} -
\score_{\pi(\upindpone{\setind}) }.
\end{align}
The dependence on these gaps is controlled via the functions
\begin{align}
\FLOW(x) \defeq \frac{1}{x^2}, \quad \mbox{and} \quad \FAR(x) \defeq
\frac{\log(2\log(2/x))}{x^2}.
\end{align}
In part (a) of the theorem to follow, we prove an upper bound
involving $\FAR$ on the AR algorithm, and in part (b), we prove a
lower bound involving $\FLOW$ that applies to \emph{any} uniformly
$\delta$-accurate algorithm.  As one might intuitively expect, the
number of comparisons required is lower when the gaps between the
underlying scores are larger. See Figure~\ref{fig:gaps} for an
illustration of the gaps for the particular problem of finding a
partitioning of the items $\{1,2,\ldots, 6\}$ into three sets of
cardinality two each.


\begin{figure}[t,b]
\begin{center}
\figuregaps
\end{center}
\captionof{figure}{\label{fig:gaps} Illustration of the gaps
  $\gapup{\setind}{i}$ and $\gaplo{\setind}{i}$ relevant for finding a
  partitioning of the items $\{1,2,\ldots,6\}$ into the sets $\S_1 =
  \{1,2\}$, $\S_2 = \{3,4\}$, and $\S_3 = \{5,6\}$.
}
\end{figure}

 
\begin{theorem}
\ \newline
\label{ThmRanking}
There are positive universal constants $(\constup, \clow)$ such that:
\begin{enumerate}[leftmargin=0.6cm]
\item[(a)] For any $\Pmat \in \comparisonclass[0]$, and any $\delta
  \in (0,0.14]$ the AR algorithm is $\delta$-accurate for $\Pmat$
    using a query size upper bounded by
\begin{subequations}
\begin{equation}
\label{eq:numtermAR}
\constup \, \log\left( \frac{\numitems}{\delta}\right) \Big \{ \sum_{i
  \in \S_1} \FAR( \gaplo{1}{i}) + \sum_{\setind = 2}^{\numsets -1}
\sum_{i \in \S_\setind} \max \Big \{ \FAR( \gaplo{\setind}{i}), \FAR(
\gapup{\setind}{i}) \Big \} + \sum_{i \in \S_\numsets} \FAR(
\gapup{\numsets}{i}) \Big \}.
\end{equation}
\item[(b)] For any $\delta \in (0, 0.14]$, consider a ranking
  algorithm that is uniformly $\delta$-accurate over
  $\comparisonclass[1/8]$.  Then when applied to a given pairwise
  comparison model $\Pmat \in \comparisonclass[3/8]$, it must make at
  least
\begin{align}
\clow \log\left(\frac{1}{2 \delta} \right) \Big \{ \sum_{i \in \S_1}
\FLOW( \gaplo{1}{i}) \!+\!  \sum_{\setind = 2}^{\numsets -1} \sum_{i
  \in \S_\setind} \max \Big \{ \FLOW( \gaplo{\setind}{i}), \FLOW(
\gapup{\setind}{i}) \Big \} \!+\!  \sum_{i \in \S_\numsets} \FLOW(
\gapup{\numsets}{i}) \Big \}
\end{align}
\end{subequations}
comparisons on average.
\end{enumerate}
\end{theorem}

Part (a) of Theorem~\ref{ThmRanking} proves that the AR algorithm is
$\delta$-accurate, and characterizes the number of comparisons
required to find a ranking as a function of the gaps between scores.
In contrast, part (b) shows that, up to logarithmic factors, the AR
algorithm is optimal, not only in a minimax sense, but in fact when
acting on any given problem instance.  The proof of part (b) involves
constructing pairs of comparison matrices that are especially hard to
distinguish, and makes use of a change of measure
lemma~\cite[Lem.~1]{kaufmann_complexity_2014} from the bandit
literature.  For the special case of top-1 identification
(corresponding to $\numsets=2$ and $\setb_1=1$),
\citet{jamieson_sparse_2015} and \citet{urvoy_generic_2013} observe
that by using the relation to multi-armed bandits discussed in
Section~\ref{sec:relationbandits}, a standard multi-armed bandit
algorithm can be applied which in turn is known to achieve the sample
complexity~\eqref{eq:numtermAR}.  Again for the special case of top-1
identification, part (b) of Theorem~\ref{ThmRanking} recovers
Theorem~1 in \cite{jamieson_sparse_2015}. 
Note that our negative result in part (b) pertains to the stochastic regime, 
where the pairwise comparison probabilities are bounded away from zero, 
and does therefore not rule out the possibility that in the regime where the 
pairwise comparison probabilities are very close to one, improvements 
in sample complexity are possible. 

In order to gain intuition on this result, in particular the
dependence on the squared gaps, it is useful to specialize to the toy
case $\numitems = 2$.  In this special case with $\numitems=2$, we
have $\score_1 = \Pmat_{12}$ and $\score_2 = \Pmat_{21} = 1 -
\Pmat_{12}$.  Thus, the ranking problem reduces to testing the
hypothesis $\{\score_1 > \score_2\}$.
One can verify that the hypothesis $\{\score_1 > \score_2\}$ is
equivalent to $\{\pmat_{12} > \frac{1}{2} \}$.
Let $\bernrv_i,i=1,\ldots,\numcmp$ be the outcomes of $\numcmp$
independent comparisons of items $1$ and $2$, that is,
$\PR{\bernrv_i=1}=\pmat_{12}$ and $\PR{\bernrv_i=0}=1-\pmat_{12}$.  A
natural test for $\{\pmat_{12} > \frac{1}{2} \}$ is to test whether
$\bernrvover >1/2$, where \mbox{$\bernrvover \defeq
  \frac{1}{\numcmp}\sum_{i=1}^\numcmp \bernrv_i$.}  Supposing without
loss of generality that $\pmat_{12} > \frac{1}{2}$, by Hoeffding's
inequality, we can upper bound the corresponding error probability as
\begin{align*}
\PR{ \bernrvover \leq 1/2 } = \PR{ \bernrvover - \pmat_{12} \leq 1/2 -
  \pmat_{12} } \leq e^{-2 \numcmp (1/2 - \pmat_{12})^2} = e^{-2
  \numcmp (\score_1 - \score_2)^2}.
\end{align*}
Thus, for $\numcmp \ge \frac{\log(1/\delta)}{2(\score_1-\score_2)^2}$
the error probability is less than $\delta$. The
bound~\eqref{eq:numtermAR} in Theorem~\ref{ThmRanking}(a) yields an
identical result up to a logarithmic factor.

More generally, testing for the inclusion $i \in \S_\setind$ amounts
to testing for $\gapup{\setind}{i} > 0$ and $\gaplo{\setind}{i} > 0$,
where $\gapup{\setind}{i} = \score_{\pi(\loindmone{\setind})} -
\score_{i}$ and $\gaplo{\setind}{i}= \score_{i} -
\score_{\pi(\upindpone{\setind})}$. These requirements provide some
intuition regarding the dependence of our bounds on the inverses of
the squared gaps.


\subsection{Gains due to active estimation}

In order to understood the benefits of an active strategy, it is
worthwhile to compare the performance of our active method to the
(minimax optimal) guarantees obtainable by passive comparison
strategies.  We hasten to add that these gains should not be seen as
surprising in of themselves, since it is well-known that active
estimators can often yield significant improvements over passive
schemes.

Recent work by a subset of the current authors~\cite{shah_simple_2015}
considers the problem of ranking items from pairwise comparisons in a
passive random design setup.  On one hand, it is shown \mbox{(Theorem
  1)} that a simple passive scheme---namely, one that ranks items
according to the total number of comparisons won---recovers the top
$k$ items with high probability using $\frac{\numitems \log \numitems
}{(\score_k - \score_{k+1})^2}$ comparisons in total; the same paper
also establishes a matching lower bound, meaning that no passive
scheme can do better up to constant factors.  In contrast,
Theorem~\ref{ThmRanking} of the present paper shows that in the active
setting, the number of comparisons necessary and sufficient for
finding the top $k$ items is of the order
\begin{align*}
\sum_{i=1}^{k} \frac{1}{ (\score_i - \score_{k+1})^{2}} +
\sum_{i=k+1}^\numitems \frac{1}{ (\score_k - \score_i)^{2}},
\end{align*}
up to a logarithmic factor.  By comparing this guarantee to the
passive sample complexity $\frac{\numitems \log \numitems }{(\score_k
  - \score_{k+1})^2}$, we can understand when active strategies do or
do not lead to substantial gains.  First, note that the complexity of
the non-active estimator is always lower, except for scores satisfying
the linear constraints $\score_1= \ldots = \score_k$ and
$\score_{k+1}= \ldots = \score_\numitems$, in which case the two
estimators would have similar performance.  Second, the difference
in sample complexity can be as large as a factor of $\numitems$, up to
logarithmic factors.  In particular, suppose that the score difference
$\score_i - \score_{i+1}$ is on the order of $1/\numitems$: in this
case, up to logarithmic factors, the sample complexity of the active
and passive schemes scale as $\numitems^2$ and $\numitems^3$
respectively.  A similar conclusion holds if we compare the results of
the paper~\cite{shah_simple_2015} with those of the present paper for
the problem of recovering the full ranking.

\begin{figure}[t]
\begin{center}
\begin{tikzpicture}
\begin{groupplot}[
         title style={at={(0.5,-0.35)},anchor=north}, group
         style={group size=3 by 1, xlabels at=edge bottom, ylabels
           at=edge left,yticklabels at=edge left,horizontal
           sep=0.2cm}, xlabel={index}, ylabel={estimated score},
         width=0.38\textwidth, ymin=0,ymax=1]
          \nextgroupplot[title= {(a) PlaNYC}] \addplot +[only
            marks,blue,mark=x] table[x index=0,y
            index=1]{./fig/PlaNYC_scores.dat};
          \nextgroupplot[title={(b) OECD}] \addplot +[only
            marks,blue,mark=x] table[x index=0,y
            index=1]{./fig/OECD_scores.dat};
\end{groupplot}          
\end{tikzpicture}
\end{center}
\caption{\label{fig:borda}
Estimated scores from comparisons of the proposals in the PlaNYC (a) and
OECD (b) surveys, as reported in the paper~\cite{salganik_wiki_2015}
(only scores of items (proposals) that were rated at least $50$ times are
depicted).  Estimation of the top $k$ proposals or another ranking with an
active scheme would require a significantly smaller number of queries
compared to a non-active estimator. }
\end{figure}
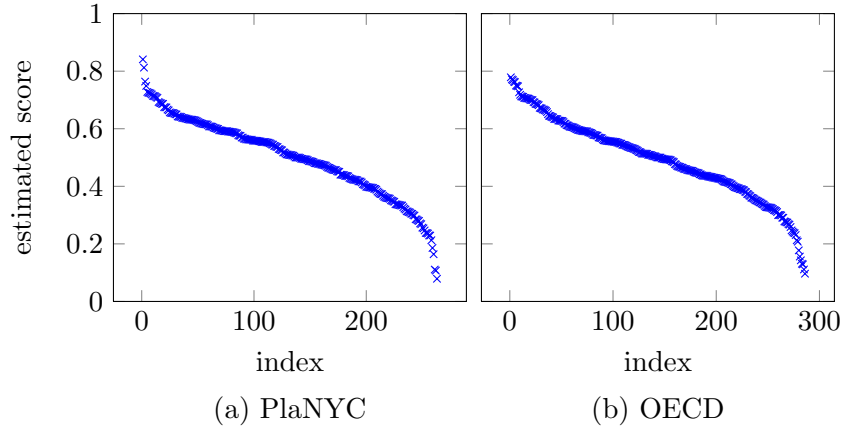

Having seen that the gains from active estimation depend on the
distribution of the scores $\SHORTSCORE$, it is natural to
wonder how these scores behave in real-world settings.  As one
illustration, Figure~\ref{fig:borda} shows some real-world examples of
this distribution for data collected by~\citet{salganik_wiki_2015};
the left panel shows the scores estimated in the
paper~\citep{salganik_wiki_2015} of a collection of environmental
proposals for New York City, whereas the right panel shows a
collection of educational proposals for the Organisation for Economic
Co-operation and Development (OECD).  These data were collected by
asking interviewees in corresponding online surveys for preferences
between two options.  The goal of such online surveys is, for example,
to identify the top proposals or a total ranking of the proposals.
Our results show that estimation of the top $k$ proposals or another
ranking with an active scheme would require a significantly smaller
number of queries compared to an non-active estimator.

\section{When parametric assumptions don't help}
\label{sec:parametric} 

The active ranking algorithm described and analyzed in the previous
section applies to \emph{any} comparison matrix $\Pmat$---that is, it
neither assumes nor exploits any particular structure in $\Pmat$, such
as that imposed by the parametric models described in
Section~\ref{sec:intro_parametric}.  Given that the AR algorithm
imposes no conditions on the model, one might suspect that when
ranking data is actually drawn from a parametric model---for example,
of BTL or Thurstone type---it could be possible to come up with
another algorithm with a lower sample complexity.  Surprisingly, as we
show in this section, this intuition turns out to be false in the
following sense: for stochastic comparison models---in which the
comparison probabilities are bounded strictly away from zero and
one---imposing parametric assumptions can lead to at most a
logarithmic reduction in sample complexity.

Recall that a parametric model is described by a continuous and
strictly increasing CDF $\MYCDF$; in this section, we prove a lower
bound that applies even to algorithms that are given \emph{a priori}
knowledge of the function $\MYCDF$.  For any pair of constants $0 <
\pdfmin \leq \pdfmax < \infty$, we say that a CDF $\MYCDF$ is
$(\pdfmin,\pdfmax,\pmatmin)$-bounded, if it is differentiable, and if
its derivative $\MYCDF'$ satisfies the bounds
\begin{align}
\label{eq:assumptionderivativeCDF}
\pdfmin \leq \MYCDF'(\timeind) \leq \pdfmax, \quad \text{for all } t
\in [\inv{\MYCDF}(\pmatmin), \inv{\MYCDF}(1-\pmatmin)].
\end{align}
 Note that these conditions hold for standard parametric models, such
 as the BTL and Thurstone models.

The following result applies to any parametric model $\parfamily$
described by a CDF of this type.  It also involves the complexity
parameter
\begin{align}
\label{def:loboSC}
\complexityP(\vscore(\Pmat)) & \defeq \sum_{i \in \S_1} \FLOW(
\gaplo{1}{i}) + \sum_{\setind = 2}^{\numsets -1} \sum_{i \in
  \S_\setind} \max \Big \{ \FLOW( \gaplo{\setind}{i}), \FLOW(
\gapup{\setind}{i}) \Big \} + \sum_{i \in \S_\numsets} \FLOW(
\gapup{\numsets}{i}),
\end{align}
which appeared previously in the lower bound from
Theorem~\ref{ThmRanking}(b).


\begin{theorem}
\ \newline
\label{thm:necessityparametric} 
  \begin{enumerate}[leftmargin=0.6cm]
  \vspace{-0.6cm}
\item[(a)] Given a tolerance $\delta \in (0, 0.15]$, and a continuous
  and strictly increasing CDF $\Phi$ whose derivative is
  $(\pdfmin,\pdfmax,\pmatmin)$-bounded, consider any algorithm 
  that is uniformly $\delta$-accurate over $\parfamily \cap
  \comparisonclass[\pmatmin]$.  Then, when applied to a given pairwise
  comparison matrix \mbox{$\Pmat \in \parfamily \cap
    \comparisonclass[\pmatmin]$}, it must make at least  
 \begin{align}
\label{EqnLowerParametric}
 \clobo \log \left(\frac{1}{2\delta}
 \right) \complexityP(\vscore(\Pmat)), \qquad \mbox{where $\clobo
   \defeq \frac{\pmatmin \pdfmin^2}{2.004\pdfmax^2}$},
 \end{align}
 comparisons on average. 
\item[(b)] Let $\vscore \in (0,1)^\numitems$ be any set of scores that
  is realizable by some pairwise comparison matrix \mbox{$\Pmat' \in
    \comparisonclass[\pmatmin]$}, $\pmatmin>0$. Then for any continuous and strictly
  increasing $\MYCDF$, there exists a pairwise comparison matrix in
  $\Pmat \in \parfamily[\pmatmin]$ with scores $\vscore$, and in
  particular with $\complexityP(\vscore(\Pmat)) =
  \complexityP(\vscore(\Pmat'))$.
\end{enumerate}
\end{theorem}

First, let us provide some concrete settings of the constant $\clobo$:
for \mbox{$\pmatmin = \frac{3}{8}$,} we have \mbox{$\clobo = 0.164$}
and \mbox{$\clobo = 0.169$} for the BTL and Thurstone models,
respectively; whereas for \mbox{$\pmatmin = \frac{1}{4}$,} we have
\mbox{$\clobo = 0.07$} and \mbox{$\clobo = 0.079$} for the BTL and
Thurstone models, respectively.
  
Second, let us turn to the implications of
Theorem~\ref{thm:necessityparametric}.  To start, it should be noted
that the lower bound~\eqref{EqnLowerParametric} is, at least in a
certain sense, stronger than the lower bound from
Theorem~\ref{ThmRanking}, because it applies to a broader class of
algorithms---namely, those that are $\delta$-accurate \emph{only} over
the smaller class of parametric models.  On the flip side, it is
possible that the lower bound~\eqref{EqnLowerParametric} could be
weaker in some sense: more precisely, could there be some
``difficult'' matrix $\Pmat' \in \comparisonclass[\pmatmin]$ such that
the supremum of $\complexityP(\vscore(\Pmat))$ over $\Pmat \in
\parfamily[\pmatmin]$ is much smaller than
$\complexityP(\vscore(\Pmat'))$?  Part (b) of the theorem rules out
this possibility: it guarantees that for any pairwise comparison
matrix $\Pmat'$---which need not be generated by a parametric
model---there exists a parametric model $\Pmat$ for which the ranking
problem is equally hard.  This result is surprising because one might
think that imposing parametric assumptions might simplify the ranking
problem.  In fact, the full set $\comparisonclass[\pmatmin]$ is
substantially larger than the parametric subclass $\parfamily \cap
\comparisonclass[\pmatmin]$; in particular, one can demonstrate
matrices in $\comparisonclass[\pmatmin]$ that cannot be
well-approximated by any parametric model; for example, see the
paper~\cite{shah_stochastically_2015} for inapproximability results of
this type.

A consequence of Theorem~\ref{thm:necessityparametric} is that up to
logarithmic factors, the AR algorithm is again optimal, even if we
restrict ourselves to algorithms that are uniformly $\delta$-accurate
\emph{only} over a parametric subclass.  Thus, for stochastic
comparison models, imposing parametric assumptions only limits the
flexibility while failing to provide any significant reductions in
sample complexity for ranking.  It is worth commenting that for
deterministic or near-deterministic comparison models---in which the
pairwise probabilities can be arbitrarily close to zero or one--- the
constant $\clobo$ in the lower bound~\eqref{EqnLowerParametric} can
become small.  For this reason, our lower bound does not contradict
the fact that parametric assumptions might help for
(near)-deterministic comparison models.  As one example, recalling
that the BTL model described in Section~\ref{sec:intro_parametric} is
based on a parameter vector $\parw \in \reals^\numitems$, suppose that
we set $\parw_i = \xi (n-i)$ for all $i \in [\numitems]$, and then let
$\xi$ tend to infinity.  Since $\pmat_{ij} =
\frac{e^{\parw_i}}{e^{\parw_i} + e^{\parw_j}}$ under the model, taking
the limit $\xi \to \infty$ leads to a fully deterministic comparison
model in which items $i$ beats $j$ with probability one if and only if
$\parw_i > \parw_j$.  In this limit, pairwise ranking reduces to a
deterministic sorting problem, and sorting-based algorithms
(e.g.,~\cite{szorenyi_online_2015}) can be used to achieve top item
identification with $O(\numitems \log \numitems )$ comparisons.  In
contrast, in this deterministic setting, the AR algorithm
requires\footnote{To be very clear, this example does \emph{not}
  violate any of our claimed results since the lower bound of
  Theorem~\ref{ThmRanking}(b), and hence the associated claim of
  optimality, applies only to the case when the pairwise comparison
  probabilities are bounded away from $0$ and $1$ by some constant $\pmatmin$.}  $O(\numitems^2
\log \numitems)$ comparisons, which can be guaranteed by applying
Theorem~\ref{ThmRanking}(a) with the associated score vector $\score_i
= 1-\frac{i-1}{\numitems-1}$.


\section{Numerical results}
\label{sec:simulations}

We now turn to some numerical comparisons of our active ranking (AR)
algorithm with algorithms designed for parametric models. One
finding---consistent with our theory---is that the AR algorithm
is on par or outperforms these algorithms, unless the pairwise comparison
probabilities are close to zero or one. Moreover, we find that
algorithms designed for parametric models start to break down even if
the parametric modeling assumption is only slightly violated.
Finally, we experiment with the choice of constants setting confidence intervals $\alpha_\timeind$ for the AR algorithm, and find that
the choice given by our theory is conservative.


\subsection{Comparison to algorithms tailored to parametric models}

Our results in Section~\ref{sec:parametric} show that for stochastic
comparison models, algorithms that exploit parametric structure can
have sample complexities lower by at most a logarithmic factor.  On
the other hand, for (near)-deterministic comparison models, we gave an
example showing that parametric structure can allow for significant
gains.  In this section, we perform some numerical experiments to
quantify and understand these two different regimes.

To this end, we consider the problem of top item recovery, known as
the dueling bandit problem, simply because algorithms are available
for this special case of the more general ranking problem considered
in our paper.  We compare the AR algorithm to the Plackett-Luce PAC
(PLPAC)~\cite{szorenyi_online_2015} and Beat the Mean Bandit
(BTMB)~\cite{yue_beat_2011} algorithms.  Both algorithms yield an
$\delta$-accurate ranking provided the BTL modeling assumptions hold.
We choose the PLPAC algorithm for comparison as it is based on
sorting: a BTL problem with pairwise comparison probabilities close to
one and zero is in essence a noisy sorting problem, thus we expect
sorting based procedures to work well here.  The BTMB algorithm is
guaranteed to succeed if Strong Stochastic Transitivity (SST) (or a
relaxed version thereof) and a certain stochastic transitivity
triangle inequality hold\footnote{A necessary and sufficient condition
  for a matrix to satisfy the SST condition is the existence of a
  permutation of the items, such that the permuted pairwise comparison
  matrix $\pmat$ is non-decreasing across rows and non-increasing
  across columns. The stochastic transitivity inequality demands that
  for each triplet with $\score_1 > \score_j > \score_k$, we have
  that~\mbox{$\pmat_{1j}-1/2 + \pmat_{jk}-1/2 \geq \pmat_{1k} -
    1/2$}.}; both assumptions are satisfied for the BTL model.
Regarding the algorithms parameters; for the AR algorithm we set
$\alpha_t = \frac{1}{4}\sqrt{\log(3 \numitems \log(1.12 t)/\delta) }$
(see Section~\ref{sec:confinterval} for a discussion for the choice of
$\alpha_t$), and for all algorithms we choose $\delta=0.1$.

We set $\numitems = 10$ and consider two different BTL models
parameterized by $\eta>0$ and $\xi>0$, respectively, and denoted by
$\Pmat^{(\eta)}$ and $\Pmat^{(\xi)}$.  The parameters $\eta$ and $\xi$
determine how close the minimal and maximal pairwise comparison
probabilities are to $0$ and $1$; the larger, the closer.
Specifically, the parameters of the BTL model $\Pmat^{(\eta)}$ are
given by $\parw_i = \log(1/\eta + \numitems - i), i =
1,\ldots,\numitems$. This results in pairwise comparison probabilities
$\pmat_{ij}^{(\eta)} = \frac{1/\eta+\numitems - i}{2(1/\eta +
  \numitems) - i - j}$.  The parameters of the second BTL model,
$\Pmat^{(\xi)}$, are $\parw_i = \xi (\numitems-i)$ which implies that
the probability that item $i$ beats the next best item $i+1$ is
$\pmat_{i,i+1}^{(\xi)} = \frac{1}{1+e^{-\xi}}$.  Thus, each item beats
all lower ranked ones with probability at least
$\frac{1}{1+e^{-\xi}}$, which results in \emph{all} the pairwise
comparison probabilities being skewed away from $1/2$; the larger
$\xi$ the ``closer'' those probabilities are to $0$ and $1$.

\begin{figure}[t]
\begin{center}
\includegraphics{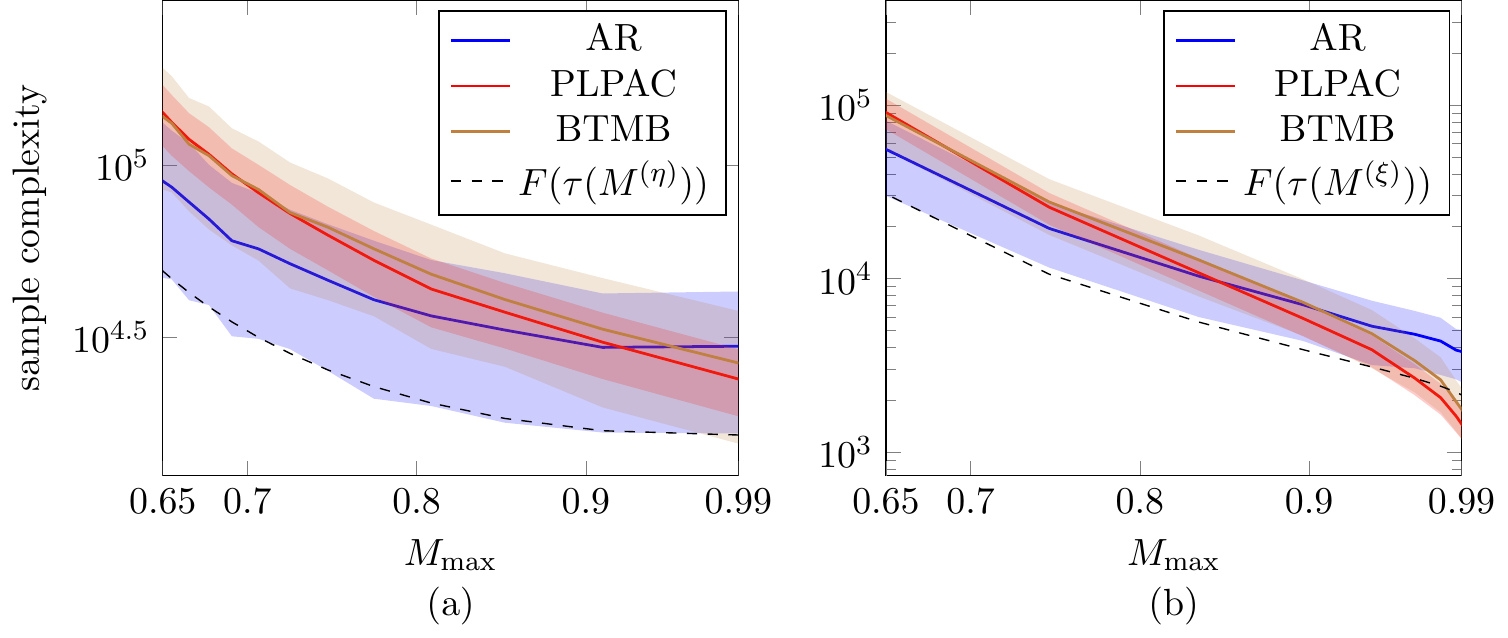}
\end{center}
\caption{\label{fig:btlcomparision} (a) Empirical sample complexity of
  the AR, PLPAC, and BTMB algorithms applied to the BTL model
  $\pmat^{(\eta)}$ with parameters $\parw_i = \log(\eta + \numitems -
  i), i = 1,\ldots,10$, and (b) applied to the BTL model
  $\pmat^{(\xi)}$ with parameters $\parw_i = \xi (\numitems-i),
  i=1,\ldots,10$, as a function of $\pmatmax \defeq \max_{i,j}
  \pmat_{ij}$. For panel (a) and (b) we varied $\eta$ and $\xi$ such
  that $\pmatmax \in [0.65,0.99]$.  The error bars correspond to one
  standard deviation from the mean.  While the AR algorithm has even
  lower sample complexity than the PLPAC and BTMB algorithms in the
  regime where $\pmatmax$ is not to close to $1$; the PLPAC and BTMB
  perform better when $\pmatmax$ is close to one.}
\end{figure}

In Figure~\ref{fig:btlcomparision} we depict the empirical sample
complexity for both models as a function of \linebreak\mbox{$\pmatmax
  \defeq \max_{i,j} \pmat_{ij}$}, along with the corresponding
complexity parameters $\complexityP( \score(\Pmat^{(\eta)}) )$ and
$\complexityP( \score(\Pmat^{(\xi)}) )$. Here, we choose the model parameters
$\eta$ and $\xi$ such that $\pmatmax$ varies between $0.65$ and
$0.99$.  The results show, as predicted by our theory, that the sample
complexity of the AR algorithm is essentially a constant times the
complexity parameter $\complexityP$. 
In contrast, the sample
complexity of the PLPAC and the BTMB algorithms improves in $\pmatmax$
relative to the complexity parameter $\complexityP$.  Note that the AR
algorithm performs better than PLPAC and BTMB if $\pmatmax$ is not too
large, while both PLPAC and BTMB have lower sample complexity than the
AR algorithm in the regime where $\pmatmax$ is very close to one.  We
remark that the relative improvement is not determined solely by
$\pmatmax$, as shown by the curves for the two differently
parameterized BTL models differing.

\begin{figure}[t]
\begin{center}
\includegraphics{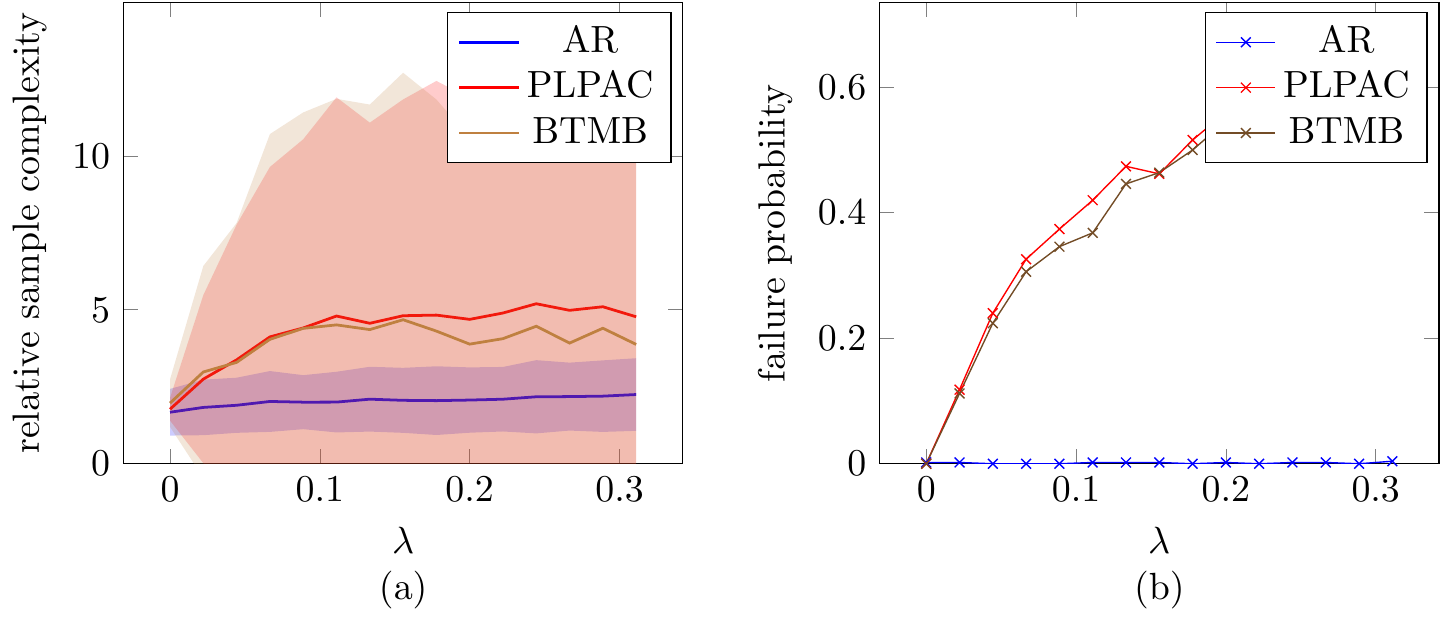}
%
%
\end{center}
\caption{\label{fig:violationBTL} (a) Relative sample complexity
  defined as the number of comparisons until termination, $\numcmp$,
  divided by the complexity parameter $\complexityP(\score(\Pmat))$,
  and (b) failure probability on a BTL model $\pmat$ with
  $\numitems=10$ and with a fraction of $\lambda$ of the off-diagonals
  of $\pmat$ substituted by a random pairwise comparison probability.
  The model transitions from a BTL model to a random pairwise
  comparison matrix in $\lambda$; the closer $\lambda$ to zero the
  closer $\pmat$ to the original BTL model.  The results show that,
  while the AR algorithm yields an $\delta$-accurate ranking after
  $O(\complexityP(\score(\Pmat)))$ comparisons, irrespectively of
  $\lambda$, the sample complexity and more importantly the failure
  probability of the PLPAC and BTMB algorithms become very large in
  $\lambda$.  }
\end{figure}

Our next simulation shows that, however, even if the pairwise
comparison matrix only deviates slightly from the BTL model, both the
sample complexity and more pertinently the failure probability (that
is, $\PR[\Pmat]{ \Shat_\setind \neq \S_\setind, \text{ for one or more
  } \setind=1, \ldots,\numsets }$) can become very large.
Specifically, as before, we generate a BTL model $\pmat$ with
$\numitems =10$ and parameters $\parw_i = \log(1 + \numitems - i), i =
1,\ldots, \numitems$.
We then substitute a fraction of $\lambda$ of the off-diagonal
elements of $\pmat$ with a number drawn uniformly from $[0,1]$.  Thus,
the model $\Pmat$ transitions from a BTL model to a random pairwise
comparison matrix in $\lambda$; for small $\lambda$, the model $\pmat$
is close to the original BTL model.  The results, depicted in
Figure~\ref{fig:violationBTL}, show that, while the AR algorithm
succeeds for all values of $\lambda$ as expected, the sample
complexity and more importantly the failure probability of the PLPAC
and BTMB algorithms become very large.  We hasten to add that both the
PLPAC and BTMB algorithm are not designed for this scenario; therefore
it might not be surprising that they fail. The
results show that these algorithms are, however, not robust to
violations of their assumed models.


\subsection{\label{sec:confinterval}Selection of confidence interval}

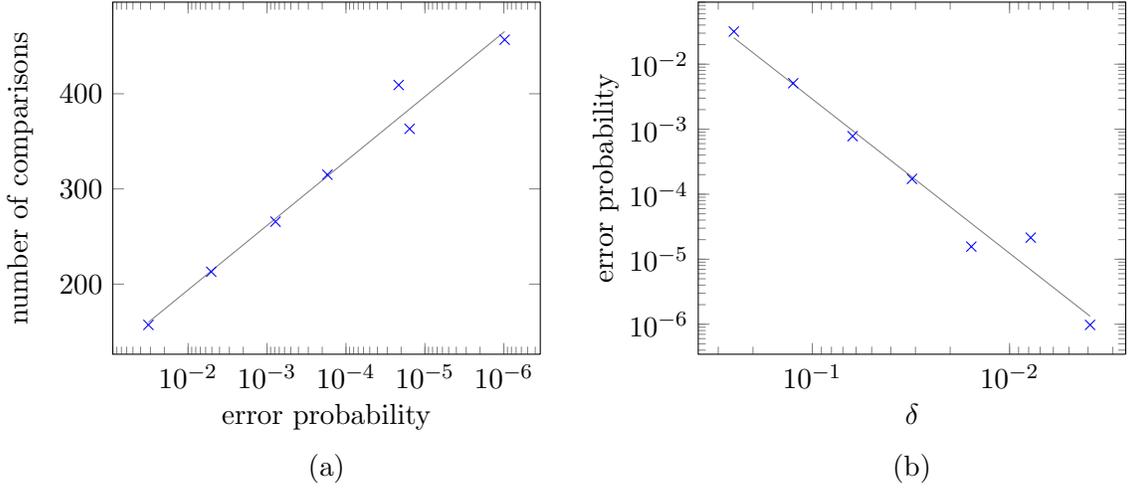
\begin{figure}[t]
\begin{center}
\begin{tikzpicture}
\begin{groupplot}[
group style={group size=3 by 1, horizontal sep=2.1cm},
width=0.44\textwidth, title style={at={(0.5,-0.3)},anchor=north},
]

\nextgroupplot[ xlabel = {error prob.}, x
  dir=reverse, ylabel = {number of comparisons}, xmode=log,x
  dir=reverse, xlabel = {num cmp}, xlabel = {error probability},legend
  pos=north west,title={(a)}] \addplot +[mark size=2.7,only
      marks,blue,mark=x] table[x index=1,y
      index=2]{./fig/cmprules_res.dat}; 
 \addplot    +[gray,solid,mark=none] table[x index=0,y
      index=1]{./fig/cmprules_fit1.dat};

\nextgroupplot[mark size=2.7,xmode =
    log,ymode=log, xlabel = {$\delta$},x dir=reverse, ylabel = {error
      probability},title={(b)}] 
\addplot +[mark
      size=2.7,only marks,blue,mark=x] table[x index=0,y
          index=1]{./fig/cmprules_res.dat};

\addplot +[gray,solid,mark=none] table[x index=0,y
          index=1]{./fig/cmprules_fit2.dat};

\end{groupplot}  
\end{tikzpicture}
\end{center}
\caption{\label{fig:cmprules} (a) Number of comparisons required to
  find the top-$2$ items out of $5$ items, for $\alpha_t = \frac{1}{4}
  \sqrt{ \log(\numitems/3 (\log(t)+1) / \delta) / t}$.  (b) Empirical
  error probability required to find the top-$2$ items out of $5$
  items.  The particular choice of the constants in $\alpha_\timeind =
  \sqrt{\log(125 \numitems \log(1.12 \timeind)/\delta) / \timeind}$ in
  our theoretical results is very conservative, in the sense that for
  obtaining a $\delta$-accurate ranking, the constants in $\alpha_t$
  can be chosen smaller, which in turn results in fewer comparisons.
}
\end{figure}

Recall that the AR algorithm eliminates an item if the confidence that
it belongs to one of the sets $\S_1,\ldots,\S_\numsets$, is
sufficiently large.  Our main results show that the AR algorithm
succeeds at recovering the ranking with probability at least
$1-\delta$, provided that the length of the confidence interval is
chosen as $\alpha_\timeind = \sqrt{\frac{\log(125 \numitems \log(1.12
    \timeind)/\delta)}{\timeind}}$.  While this result is optimal up
to log-factors, the particular choice of the constants might be overly
conservative, and improvements in the (empirical) sample complexity
might be obtained by choosing the constants in $\alpha_t$ smaller, as
we show next.  To investigate this claim, we set $\alpha_t =
\frac{1}{4} \sqrt{ \frac{ \log(\numitems/3 (\log(t)+1) /
    \delta)}{t}}$.  We generate a pairwise comparison model with
\mbox{$\numitems = 5$,} scores \mbox{$\vscore = (0.9, 0.7, 0.5, 0.3,
  0.1)$,} and use the AR algorithm to find the top $2$ items, for
different values of the desired accuracy $\delta$.  The results,
depicted in Figure~\ref{fig:cmprules}, show that, even with those
significantly smaller constants, the AR algorithm is
$\delta$-accurate.

\section{\label{sec:proofs}Proofs}

In this section, we provide the proofs of our two main theorems.  In
order to simplify notation, we take the underlying permutation $\pi$
equal to the identity, so that $\score_{1} > \score_{2} > \ldots >
\score_{\numitems}$.  This assumption entails no loss of generality,
since it can also be satisfied by re-indexing the items if necessary.

\subsection{Proof of Theorem~\ref{ThmRanking}(a)}

\newcommand\eventscore{\event_\alpha} 

In this section, we provide a proof of the achievable result stated in
part (a) of Theorem~\ref{ThmRanking}.  Our proof consists of three
main steps.  We begin by showing that the estimate
$\scorehat_i(\timeind)$ is guaranteed to be $\alpha_\timeind$-close to
$\score_i$, for all $i \in \S$, with high probability.  We then use
this result to show that the AR algorithm never misclassifies any
item, and that it stops with the number of comparisons satisfying the
claimed upper bound.

Throughout the paper, we use $\S$ to denote the set of items that have
not been ranked yet; to be clear, since items are eliminated from $\S$
at certain time steps $\timeind$, the set $\S$ changes with
$\timeind$, but we suppress this dependence for notational simplicity.

\begin{lemma}
\label{lem:probound}
Under the theorem's assumptions, the event
\begin{align}
\label{eq:taubounded}
\eventscore \defeq \{ \left| \scorehat_i(\timeind) - \score_i \right|
\leq \alpha_\timeind, \quad \text{for all $i \in \S$ and for all
  $\timeind \geq 1$} \}
\end{align}
occurs with probability at least $1 - \delta$.
\end{lemma}
 
Our next step is to show that provided that the event $\eventscore$
occurs, the AR algorithm never misclassifies any item, that is,
$\Shat_\setind \subseteq \S_\setind$ for all $\ell$ and for all $t\geq
1$.  First suppose that, at a given time step $\timeind$, the AR
algorithm did not misclassify any item at a previous time step.  We
show that, at time $\timeind$, conditioned on the event $\eventscore$,
any item $j \in \mc \S$ is added to $\Shat_\setind$ only if $j \in
\S_\setind$, which implies that the AR algorithm does not misclassify
any item at time $t$.  This fact is a consequence of our second
auxiliary result.

In order to state this second lemma, we require some additional
notation. Let $\score_{\SPECSCORE{k}}$ denote the $k$-th largest score
among the \emph{latent} scores $\score_i,\ i \in \S$. Note that we use
the notation $\SPECSCORE{\cdot}$ to emphasize that the index
$\SPECSCORE{k}$ is not necessarily equal to the index $(k)$, since the
latter corresponds to the $k$-th largest score amongst the
\emph{estimated} scores $\scorehat_i(\timeind) ,\ i\in \S$.

\begin{lemma}
\label{lem:noerror}
Suppose that the event $\eventscore$ occurs.  Then both of the
implications
\begin{subequations}
\begin{align}
\bullet \quad &\text{for any $j \in \S$, \quad
    $\scorehat_{j}(\timeind) < \scorehat_{(
      \ARloindmone{\setind})}(\timeind) - 4 \alpha_\timeind$\quad
    implies \quad $\score_j <
    \score_{\SPECSCORE{\ARloindmone{\setind}}}$}, \text{ and }
\label{eq:implication1} \\
\bullet \quad &\text{for any $j \in \S$,\quad $\scorehat_{j}(\timeind) >
  \scorehat_{(\ARupindpone{\setind})}(\timeind) + 4 \alpha_\timeind$
  \quad implies \quad $\score_j >
  \score_{\SPECSCORE{\ARupindpone{\setind}}}$, }
\label{eq:implication2}
\end{align}
\end{subequations}
hold for all $t \geq 1$.
\end{lemma}
Provided that the AR algorithm did not misclassify any item at a
previous time step, some consequences of
implications~\eqref{eq:implication1} and~\eqref{eq:implication2} are
the following: 
\begin{itemize}
\item first, for any index $\ell$, an item is added to $\Shat_\setind$ at time
$\timeind$ only if $j \in \S_\setind$.
\item therefore, we are guaranteed that $\Shat_\setind \subseteq
  \S_\setind$ at time $\timeind + 1$.
\end{itemize}
These consequences allow us to apply an inductive argument to conclude
that the AR algorithm never misclassifies any item.\\

Our next step is to show that, conditioned on the event $\eventscore$
on which the AR algorithm does not misclassify any item, all items are
eliminated after the number of comparisons given in
equation~\eqref{eq:numtermAR} have been carried out.  Since, by
Lemma~\ref{lem:probound}, the event $\eventscore$ holds with
probability at least $1 - \delta$, this concludes the proof of
Theorem~\ref{ThmRanking}(a).

In order to establish the former claim, we use the following lemma, in
which we made the dependence of the set of candidates $\S$ on
$\timeind$ explicit by writing $\S(\timeind)$.

\newcommand\constone{c_1}

\begin{lemma}
\label{lem:sampcomp}
Suppose that the event $\eventscore$ occurs.  For any index $\setind
\in \{2,\ldots,\numsets\}$ and any item $i \in \S_\setind \cap
\S(\timeup_i)$, we have, with $\constone \defeq 654$,
\begin{subequations}
\begin{align}
\scorehat_i(\timeup_i) <
\scorehat_{(\ARloindmone{\setind})}(\timeup_i) - 4 \alpha_{\timeup_i},
\quad \mbox{where~} \timeup_i
\defeq \frac{\constone}{ \gapup{\setind}{i}^2 } \log\left(
\frac{n}{\delta} \log\left(\frac{2}{\gapup{\setind}{i}} \right)
\right) , \quad \gapup{\setind}{i} = \score_{\loindmone{\setind}} -
\score_i,
\label{eq:tiup}
\end{align}
and for $\ell \in \{1,\ldots,\numsets-1\}$ and any item $i \in
\S_\setind \cap \S(\timelo_i)$, with probability at least $1-
\frac{\delta}{4\numitems}$, we have
 \begin{align}
\scorehat_i(\timelo_i) >
\scorehat_{(\ARupindpone{\setind})}(\timelo_i) - 4 \alpha_{\timelo_i},
\quad \mbox{where~} \timelo_i \defeq \frac{\constone}{
  \gapup{\setind}{i}^2 } \log\left( \frac{n}{\delta}
\log\left(\frac{2}{\gaplo{\setind}{i}} \right) \right) , \quad
\gaplo{\setind}{i} = \score_i - \score_{\upindpone{\setind}}.
\label{eq:tilo}
\end{align}
\end{subequations}
\end{lemma}


Consequently, the index $i \in \S_\setind$ is eliminated from the set
of candidates $\S$ after no more than the following number of many
time steps (and hence comparisons):
\begin{align*}
\begin{cases}
\timelo_i, &\text{if } \setind = 1 \\ \max(\timelo_i,\timeup_i),
&\text{if } \setind \in \{2, \ldots, \numsets-1\} \\ \timeup_i,
&\text{if } \setind = \numsets
\end{cases}.
\end{align*}
 Using the relations
\begin{align*}
\timeup_i \leq \constup \frac{\log(2\log(2/ \gapup{\setind}{i})) }{
  \gapup{\setind}{i}^2} \log(n/ \delta), \quad \text{and}\quad
\timelo_i \leq \constup \frac{\log(2\log(2/ \gaplo{\setind}{i})) }{
  \gaplo{\setind}{i}^2} \log(n/ \delta),
\end{align*}
where the inequalities hold for some constant $\constup$, it follows
that the AR algorithm terminates after the number of comparisons
stated in equation~\eqref{eq:numtermAR} has been carried out. \\

\noindent It remains to prove Lemmas~\ref{lem:probound},
\ref{lem:noerror}, and~\ref{lem:sampcomp}, and we do so in the
following subsections.


\subsubsection{Proof of Lemma~\ref{lem:probound}}

In order to show that the event $\eventscore$ occurs with probability
at least $1-\delta$, first recall that comparing item $i$ to an item
chosen uniformly at random from $[\numitems]\backslash \{i\}$ is
equivalent to taking an independent draw from a Bernoulli random
variable with mean $\score_i$.  One can verify from the
recursion~\eqref{eq:scorehatupdate} that $\scorehat_i(t)$ is a sum of
$t$ independent Bernoulli random variables, each of which has mean
$\tau_i/\timeind$.  In order to control the fluctuations of
$\scorehat_i(t)$, we make use of a non-asymptotic version of the law
of the iterated logarithm from~\citet{jamieson_lil_2014}. 
\begin{lemma}[{\cite[Lem.~1, with $\epsilon=0.1151$]{jamieson_lil_2014}}]
\label{lem:lilUCB}
Given an i.i.d. sequence $\{\bernrv_s\}_{s=1}^\infty$ of Bernoulli
variables with mean $\mu$, then for any $\epsilon \in (0,1)$ and
$\delta' \in (0,1)$, we have
\begin{align*}
\left| \frac{1}{t} \sum_{s=1}^t (\bernrv_s-\mu) \right| \leq
\sqrt{\frac{\log(125 \log(1.12 \timeind)/\delta')}{\timeind}}, \qquad
\mbox{for all $t \geq 1$}
\end{align*}
with probability at least $1-\delta$.
\end{lemma}

\noindent In the current context, applying Lemma~\ref{lem:lilUCB} with
$\delta'=\numitems/\delta$ and $ \alpha_\timeind =
\sqrt{\frac{\log(125 \numitems \log(1.12 \timeind)/\delta)}{\timeind}}
$ yields
\begin{align*}
\mprob \Big[ \left| \scorehat_i(t) - \score_i \right| \geq
  \alpha_\timeind \text{~~~for some $\timeind\geq 1$} \Big] & \leq
\frac{\delta}{\numitems}.
\end{align*}
Taking the union bound over all indices $i\in \S \subseteq
[\numitems]$ yields that $\mprob[\eventscore] \geq 1- \delta$, as
claimed.


\subsubsection{Proof of Lemma~\ref{lem:noerror}}

We show that implications~\eqref{eq:implication1}
and~\eqref{eq:implication2} follow from the inequality in event
$\eventscore$.  In order to do so, consider any index $k'$ such that
$\scorehat_{k'}(\timeind) = \scorehat_{(k)}(\timeind)$.  Here we have
allowed for the possibility that $(k)$ may not be unique.  We start by
showing that the inequality in event $\eventscore$ implies that
\begin{align}
\label{eq:tauktaukprime}
|\score_{\SPECSCORE{k}} - \score_{k'} | \leq 2 \alpha_\timeind.
\end{align} 
We claim that $\score_{\SPECSCORE{k}} - \score_{k'} \geq - 2
\alpha_\timeind$.  By definition of $k'$, there are $k$ indices
$\{i_1,\ldots,i_k \}$ such that $\scorehat_{i_\ell}(\timeind) \geq
\scorehat_{k'}(\timeind)$ for every $\ell \in [k]$. In
conjunction with the inequality in event $\eventscore$, we obtain
\begin{align*}
\score_{i_\ell} + \alpha_\timeind \geq \score_{k'} - \alpha_\timeind.
\end{align*}
Since this inequality holds for $k$ many indices $\{i_1,\ldots,i_k
\}$, one of those indices must be $\SPECSCORE{k}$, due to 
$\score_{\SPECSCORE{1}} \geq \score_{\SPECSCORE{2}} \geq \ldots \geq \score_{\SPECSCORE{|\S|}}$. 
It follows that
$\score_{\SPECSCORE{k}} - \score_{k'} \geq -2 \alpha_\timeind$.  It
remains to establish that $\score_{\SPECSCORE{k}} - \score_{k'} \leq 2
\alpha_\timeind$.  By definition of $k'$, there are $|\S| - k +1$ many
indices $\tilde k$ obeying
\begin{align*}
\scorehat_{\tilde k}(\timeind) \leq \scorehat_{k'}(\timeind).
\end{align*}
By the inequality in event $\eventscore$, this yields $\score_{\tilde
  k} - \alpha_\timeind \leq \score_{k'} + \alpha_\timeind$.  Since
this inequality holds for $|\S|-k+1$ indices $\tilde k$, it must hold
for $\tilde k=\SPECSCORE{k}$, which implies $\score_{\SPECSCORE{k}}
\leq \score_{k'} + 2\alpha_\timeind$.  Thus, we have established that
inequality~\eqref{eq:tauktaukprime} must hold under event
$\eventscore$.

We are now ready to establish the claim~\eqref{eq:implication1}.  Let
$k'$ be any index such that $\scorehat_{k'} =
\scorehat_{(\ARloindmone{\setind})}$.  As long as
$\scorehat_{j}(\timeind) < \scorehat_{(\ARloindmone{\setind})} - 4
\alpha_\timeind$, we have
\begin{align}
4 \alpha_\timeind \; < \; \scorehat_{k'}(\timeind) -
\scorehat_j(\timeind) & \mystackrel{(i)}{\leq} \score_{k'} +
\alpha_\timeind - \score_j + \alpha_\timeind \nonumber \\
\label{eq:usetauktaukprime}
& \leq \score_{\SPECSCORE{\ARloindmone{\setind}}} - \score_j + 2
\alpha_\timeind + 2 \alpha_\timeind.
\end{align}
Here step~(i) follows by the inequality in event $\eventscore$,
whereas inequality~\eqref{eq:usetauktaukprime} follows by
inequality~\eqref{eq:tauktaukprime}. Noting that
inequality~\eqref{eq:usetauktaukprime} is equivalent to $\score_j <
\score_{\SPECSCORE{\ARloindmone{\setind}}}$, we have established the
claim~\eqref{eq:implication1}. The proof of
claim~\eqref{eq:implication2} is analogous, so  we omit the
details.


\subsubsection{Proof of Lemma~\ref{lem:sampcomp}}

We first prove that, if the event $\eventscore$ occurs, then for any given index $i \in \S_\setind$, and for all 
$\setind > 1$, inequality~\eqref{eq:tiup} holds.  Let $k'$ be any index
satisfying the equality $\scorehat_{k'}(\timeind) =
\scorehat_{(\ARloindmone{\setind})}(\timeind)$, and recall that
$\score_{\SPECSCORE{k}}$ is the $k$-th largest score out of the latent
scores $\score_i, i \in \S$. 
On the event $\eventscore$, we have
\begin{align}
\scorehat_{k'}(\timeup_i) \; \geq \; \score_{k'} - \alpha_{\timeup_i}
-\scorehat_{i}(\timeup_i) & \mystackrel{(i)}{\geq} \score_{
  \SPECSCORE{\ARloindmone{\setind}} } - 3\alpha_{\timeup_i} \nonumber \\
& \mystackrel{(ii)}{\geq} \score_{\loindmone{\setind}} - 3
\alpha_{\timeup_i} \nonumber \\
& = \gapup{\setind}{i} - 3\alpha_{\timeup_i} + \score_i \nonumber \\
&\mystackrel{(iii)}{\geq}  
\gapup{\setind}{i} - 4\alpha_{\timeup_i} + \scorehat_i(\timeup_i).
\label{eq:hattktiti}
\end{align}
Here step (i) follows by inequality~\eqref{eq:tauktaukprime}, which
holds on the event $\eventscore$, and inequality~(ii) follows from
$\score_{ \SPECSCORE{\ARloindmone{\setind}} } \geq \score_{
  \loindmone{\setind} }$, which is seen as follows.  As shown above,
on the event $\eventscore$, the AR algorithm never misclassifies any
item, therefore the $\ARloindmone{\setind}$-th largest score among the
items in the set $\S$ must be larger or equal to the
$\loindmone{\setind}$-th largest score among all scores.  Finally,
inequality (iii) follows from $\scorehat_i(\timeup_i) - \score_i \leq
\alpha_{\timeup_i}$, which holds on the event $\eventscore$.

From the definition of $\alpha_{\timeup_i}$, some algebra leads to the
lower bound
\begin{align}
\label{EqnEspressoForNihar}
\gapup{\setind}{i} > 8\alpha_{\timeup_i}.
\end{align}
See the end of this subsection for details of this calculation.
Application of inequality~\eqref{EqnEspressoForNihar} on the RHS of
inequality~\eqref{eq:hattktiti} yields
\begin{align*}
\scorehat_{k'}(\timeup_i) > 4\alpha_{\timeup_i} +
\scorehat_i(\timeup_i),
\end{align*}
which concludes the proof of inequality~\eqref{eq:tiup}.  Analogously,
it follows that inequality~\eqref{eq:tilo} holds for a given item $i$
if the event $\eventscore$ occurs.


\paragraph{Proof of the lower bound~\eqref{EqnEspressoForNihar}:}

By definition of $\alpha_\timeind$ and $\timeup_i$, we have that
\begin{align*}
\alpha_{\timeup_i}^2 & = \frac{\log\left( \frac{125 \numitems}{\delta}
  \log\left[1.12 \frac{\constone}{ \gapup{\setind}{i}^2 } \log\left(
    \frac{n}{\delta} \log\left(\frac{2}{\gapup{\setind}{i}} \right)
    \right) \right] \right)}{ \frac{\constone}{ \gapup{\setind}{i}^2 }
  \log\left( \frac{n}{\delta} \log
  \left(\frac{2}{\gapup{\setind}{i}}\right) \right) } \\
& \leq \frac{ \gapup{\setind}{i}^2 }{\constone} \frac{\log\left(
  \frac{125 \numitems}{\delta} \log\left[ \frac{ 1.12\constone }{4}
    \frac{\numitems}{\delta} \frac{2^3}{\gapup{\setind}{i}^3} \right]
  \right)}{ \log\left( \frac{n}{\delta} \log
  \left(\frac{2}{\gapup{\setind}{i}}\right) \right) } \\
& \leq \frac{ \gapup{\setind}{i}^2 }{\constone} \frac{\log\left(
  \frac{125 \numitems}{\delta} \left( \frac{ 1.12\constone }{4}
  \frac{\numitems}{\delta} \right)^{1/3} 3 \log\left(
  \frac{2}{\gapup{\setind}{i}} \right) \right)}{ \log\left(
  \frac{n}{\delta} \log \left(\frac{2}{\gapup{\setind}{i}}\right)
  \right) } \\
& \leq \frac{ \gapup{\setind}{i}^2 }{\constone} \frac{\frac{4}{3}
  \log\left( 375 \left( \frac{ 1.12\constone }{4} \right)^{1/3}
  \frac{\numitems}{\delta} \log\left( \frac{2}{\gapup{\setind}{i}}
  \right) \right)}{ \log\left( \frac{n}{\delta} \log
  \left(\frac{2}{\gapup{\setind}{i}}\right) \right), } \\
& <\frac{ \gapup{\setind}{i}^2 }{8^2},
\end{align*}
where the last inequality holds for $\constone = 654$. 


\subsection{Proof of Theorem~\ref{ThmRanking}(b)}
\label{sec:proofnecessity}

We now turn to the proof of the lower bound from
Theorem~\ref{ThmRanking}.  We first introduce some notation required
to state a useful lemma~\cite[Lem.~1]{kaufmann_complexity_2014} from
the bandit literature. 
Let $\nu = \{\nu_j\}_{j=1}^m$ be a collection
of $m$ probability distributions, each supported on the real line
$\reals$.  Consider an algorithm $\alg$, that, at times
$\timeind=1,2,\ldots$, selects the index $\ind_\timeind \in [m]$ and
receives an independent draw $\bernrv_\timeind$ from the distribution
$\nu_{\ind_\timeind}$ in response.  Algorithm $\alg$ may select
$\ind_\timeind$ only based on past observations, that is,
$\ind_\timeind$ is $\mc F_{\timeind-1}$ measurable, where
$\sigmaalgebra_\timeind$ is the $\sigma$-algebra generated by
$\ind_1,\bernrv_{\ind_1},\ldots,\ind_\timeind,\bernrv_{\ind_\timeind}$.
Algorithm $\alg$ has a stopping rule $\stoptime$ that determines the
termination of $\alg$.  We assume that $\stoptime$ is a stopping time
measurable with respect to $\sigmaalgebra_\timeind$ and obeying
$\PR{\stoptime < \infty} =1$.

Let $\numcmp_i(\stoptime)$ denote the total number of times index $i$
has been selected by the algorithm $\alg$ (until termination).  For
any pair of distributions $\nu$ and $\nu'$, we let $\mathrm{KL}(\nu,
\nu')$ denote their Kullback-Leibler divergence, and for any $p, q \in
    [0,1]$, let $\kl(p,q) \defeq p \log \frac{p}{q} + (1-p) \log
    \frac{1-p}{1-q}$ denote the Kullback-Leiber divergence between two
    binary random variables with success probabilities $p, q$.

With this notation, the following lemma relates the cumulative number
of comparisons to the uncertainty between the actual distribution
$\nu$ and an alternative distribution $\nu'$.

\begin{lemma}[{\cite[Lem.~1]{kaufmann_complexity_2014}}]
Let $\nu,\nu'$ be two collections of $m$ probability distributions on
$\reals$.  Then for any event $\event \in \sigmaalgebra_\stoptime$
with $\PR[\nu]{\event} \in (0,1)$, we have
\begin{align}
\sum_{i=1}^m \EX[\nu]{\numcmp_i(\stoptime)} \mathrm{KL}(\nu_i,
\nu_i') \geq \kl(\PR[\nu]{\event}, \PR[\nu']{\event}).
\end{align}
\label{lem:changemeasure}
\end{lemma}

\noindent Let us now use Lemma~\ref{lem:changemeasure} to prove
Theorem~\ref{ThmRanking}(b).  In particular, we apply it using the
event
\begin{align}
\label{eq:performancecriterioninnecessity}
\event & \defeq \Big \{ \Shat_\setind = \S_\setind, \quad \text{for all }
\setind =1,\ldots,\numsets \Big \},
\end{align}
which corresponds to success of the algorithm $\alg$.  Recalling that
$\stoptime$ is the stopping rule of algorithm $\alg$, we are
guaranteed that $\event \in \sigmaalgebra_\stoptime$.  Given the
linear relations $\pmat_{ij} = 1- \pmat_{ji}$, the pairwise comparison
matrix $\Pmat$ is determined by the entries $\{ \pmat_{ij},
i=1,\ldots,\numitems,\; j = i+1,\ldots,\numitems \}$.  Let
$\numcmp_{ij}(\stoptime)$ be the total number of comparisons between
items $i$ and $j$ made by $\alg$.  For any other pairwise comparison
matrix $\Pmat' \in \comparisonclass[0]$, Lemma~\ref{lem:changemeasure}
ensures that
\begin{align}
\label{eq:lemspecialized}
\sum_{i=1}^\numitems \sum_{j=i+1}^\numitems \EX[\Pmat]{\numcmp_{ij}}
\kl(\pmat_{ij}, \pmat_{ij}') \geq \kl(\PR[\Pmat]{\event},
\PR[\Pmat']{\event}).
\end{align}
For some $\setind > 1$ and item\footnote{It is helpful here to make
  explicit the dependence of $\S_\setind = \{\pi(\loind{\setind}),
  \ldots, \pi(\upind{\setind}) \}$ on the distribution $\Pmat$. Note
  that $\pi$ is a function of $\Pmat$.}  $\m \in \S_\setind(\Pmat)$,
our next step is to construct a matrix $\Pmat' \in
\comparisonclass[1/8]$ such that $\m \notin \S_\setind(\Pmat')$ under
the distribution $\Pmat'$.  Since the algorithm $\alg$ is uniformly
$\delta$-accurate over $\comparisonclass[1/8]$ by assumption, we are
guaranteed that
\begin{align*}
\PR[\Pmat]{\event} \geq 1 - \delta \quad \mbox{and} \quad
\PR[\Pmat']{\event} \leq \delta,
\end{align*}
from which it follows that
\begin{align}
\kl(\PR[\Pmat]{\event}, \PR[\Pmat']{\event}) \geq \kl(\delta,1-\delta)
= (1-2\delta) \log \frac{1-\delta}{\delta} \geq \log
\frac{1}{2\delta},
\label{eq:lbdApl}
\end{align}
where the last inequality holds for $\delta \leq 0.15$.

\begin{figure}
\begin{center}
\newcommand{\myunit}{0.7 cm} \tikzset{ node style
  sp/.style={draw,circle,minimum size=\myunit}, node style
  ge/.style={circle,minimum size=\myunit}, node style
  re/.style={draw,rectangle,minimum size=\myunit}, }

\begin{tikzpicture}

\begin{scope}[inner sep=0.04cm]

\matrix (A) [matrix of math nodes,%
             nodes = {node style ge},%
             left delimiter  = (,%
             right delimiter = )] at (0,0)
{%
& & & \node[node style sp] {\pmat_{14}}; & & \\
& & & \node[node style sp] {\pmat_{24}}; & & \\ 
& & & \node[node style sp] {\pmat_{34}}; & & \\
\node[node style re] {\pmat_{41}}; & \node[node style re]{\pmat_{42}}; & \node[node style re] {\pmat_{43}}; & \pmat_{44} & \node[node style re] {\pmat_{45}}; &  \node[node style re] {\pmat_{46}}; \\
& & & \node[node style sp] {\pmat_{54}}; & & \\
& & & \node[node style sp] {\pmat_{64}}; & & \\
};
\end{scope}
\begin{scope}[xshift=6cm,yshift=-2cm]
        \begin{axis}[xlabel={index $j$},ylabel={scores $\score_j, \score_j'$},width=6.5cm]
        \addplot[only marks,mark=o,blue] coordinates {
        (1,0.75)
        (2,0.65)
        (3,0.55)
        (4,0.45)
        (5,0.35)
        (6,0.25)
        };
	\addlegendentry{$\score_j$}
	\addplot[only marks,mark=asterisk,red] coordinates {
        (1,0.7)
        (2,0.6)
        (3,0.5)
        (4,0.66)
        (5,0.3)
        (6,0.2)
        };
	\addlegendentry{$\score_j'$}
	\addplot[dashed] coordinates {
        (1,0.625)
        (6,0.625)
        };
\end{axis}
\end{scope}
\end{tikzpicture}

\end{center}
\caption{
\label{fig:illustration}
Illustration of the distributions $\pmat,\pmat'$ and the corresponding
scores $\score_j,\score'_j$: Suppose that $\m=4$ and $\loindmonemh =
2$. The probabilities $\pmat_{ij}'$ are obtained from the
probabilities $\pmat_{ij}$ by increasing the probabilities surrounded
by a rectangle, and decreasing the probabilities surrounded by a
circle, all others remain unchanged.
}
\end{figure}
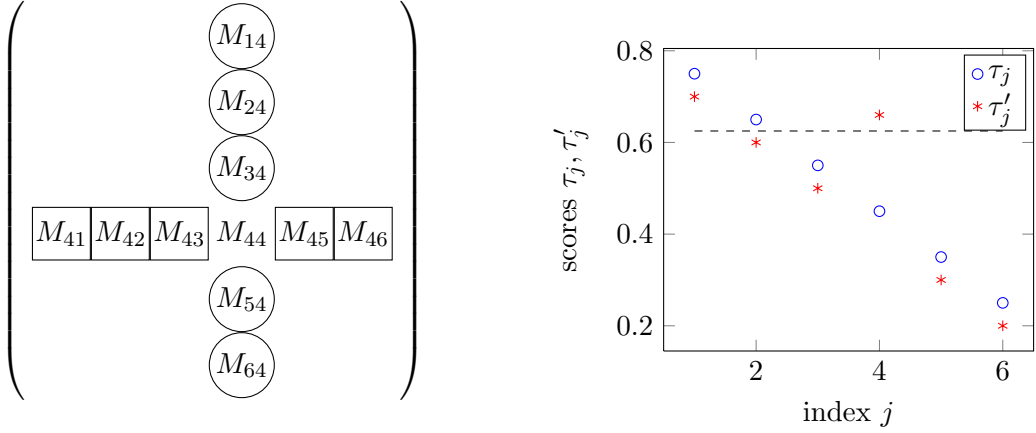

It remains to specify the alternative matrix $\Pmat' \in
\comparisonclass[0]$ for use in inequality~\eqref{eq:lbdApl}: it is
defined with entries
\begin{align}
\label{eq:pdashij}
\pmat_{ij}' & \defeq
\begin{cases}
\pmat_{\m j} + (\score_{\loindmonemh} - \score_\m ) , & \text{if }
i=\m, j \in [\numitems] \setminus \{ \m \} \\ \pmat_{i \m} -
(\score_{\loindmonemh} - \score_\m ) , & \text{if } j=\m, i \in
[\numitems] \setminus \{ \m \} \\ \pmat_{ij} & \text{otherwise}.
\end{cases}
\end{align}
From this definition, it follows that
\begin{align*}
\score_\m' = \frac{1}{\numitems-1} \sum_{j \in [\numitems] \setminus
  \{\m\}} \pmat_{\m j}' \nonumber &= \frac{1}{\numitems-1}\sum_{j \in
  [\numitems] \setminus \{\m\}} \left( \pmat_{\m j} +
(\score_{\loindmonemh} - \score_\m ) \right) \; = \;
\score_{\loindmonemh}.
\end{align*}
Similarly, all other scores $\score_i'$ are smaller than $\score_i$ by
a common constant, that is, for \mbox{$i \in [\numitems] \setminus \{ \m
  \}$}
\begin{align*}
\score_i' = \score_i - \frac{1}{n-1} (\score_{\loindmonemh} -
\score_\m ).
\end{align*}
See Figure~\ref{fig:illustration} for an illustration. 
 It follows that, under the distribution $\Pmat'$, the score of item
 $\m$ is among the $\loindmonemh$ highest scoring items, which ensures
 $\m \notin \S_\setind(\Pmat')$.  Moreover, we claim that $\Pmat' \in
 \comparisonclass[1/8]$.  This inclusion follows from the assumption
 $\Pmat \in \comparisonclass[3/8]$, which implies that
\begin{align*}
\pmat_{\m j}' \leq \frac{5}{8} + \left(\frac{5}{8} - \frac{3}{8}
\right) \leq \frac{7}{8}.
\end{align*}
An analogous argument shows that $\pmat_{\m j}' \geq \frac{1}{8}$.


Next consider the total number of comparisons of item $\m$ with all
others items,\linebreak \mbox{that is, $ \numcmp_\m = \sum_{ j \in [\numitems]
    \setminus \{\m\} } \numcmp_{\m j}$.} By the linearity of
expectation, we have
\begin{align*}
\max_{j \in [\numitems] \setminus \{\m \}} \kl(\pmat_{\m j}, \pmat_{\m
  j}') \EX[\Pmat]{\numcmp_\m} 
  &= \max_{j \in [\numitems] \setminus \{
  \m\}} \kl(\pmat_{\m j}, \pmat_{\m j}') \sum_{j' \in [\numitems]
  \setminus \{\m\} } \EX[\Pmat]{\numcmp_{\m j'}} \nonumber \\
& \geq \sum_{j \in [\numitems]\setminus\{\m \}} \EX[\Pmat]{\numcmp_{\m
    j}} \kl(\pmat_{\m j}, \pmat_{\m j}').
\end{align*}
Now observe that by the definition of $\pmat'$ in
equation~\eqref{eq:pdashij}, we have $\kl(\pmat_{ij}, \pmat_{ij}')=0$
for all $(i,j)$ outside of the sets $\{(\m,j) \mid j \in
[\numitems]\setminus \{ \m \} \}$ and $\{ (i,\m) \mid i \in
[\numitems]\setminus \{ \m \} \}$.  Removing these terms from
the sum yields
\begin{align}
\max_{j \in [\numitems] \setminus \{\m \}} \kl(\pmat_{\m j}, \pmat_{\m
  j}') \EX[\Pmat]{\numcmp_\m} & \geq \sum_{i=1}^\numitems
\sum_{j=i+1}^\numitems \EX[\Pmat]{\numcmp_{ij}} \kl(\pmat_{ij},
\pmat_{ij}') \nonumber \\ 
& \mystackrel{(i)}{\geq} \kl(\PR[\Pmat]{\event}, \PR[\Pmat']{\event})
\nonumber \\
\label{eq:bylbdApl}
& \mystackrel{(ii)}{\geq} \log \frac{1}{2\delta},
\end{align}
where step (i) follows inequality~\eqref{eq:lemspecialized} in
Lemma~\ref{lem:changemeasure}; and step (ii) follows from
inequality~\eqref{eq:lbdApl}.

We next upper bound the KL divergence on the left hand side of
inequality~\eqref{eq:bylbdApl}.  Using the inequality $\log x \leq
x-1$ valid for $x>0$, we have
\begin{align}
\kl(\pmat_{\m j}, \pmat_{\m j}')
& \leq \frac{(\pmat_{\m j} - \pmat_{\m j}')^2}{\pmat_{\m j}'
  (1-\pmat_{\m j}') } \; \leq 16 (\score_{\loindmonemh} -
\score_\m)^2,
\label{eq:klbscore}
\end{align}
where the last step uses the definition of $\Pmat'$ in
equation~\eqref{eq:pdashij}, as well as the inclusion $\frac{1}{8}
\leq \pmat_{\m j}' \leq \frac{7}{8}$, which implies that
$\frac{1}{\pmat_{\m j}' (1-\pmat_{\m j}') } \leq 16$.

Applying inequality~\eqref{eq:klbscore} to the left hand side of
inequality~\eqref{eq:bylbdApl} yields 
\begin{align}
\EX[\Pmat]{\numcmp_\m} \geq \frac{\log(1/(2\delta))}{16
  (\score_{\loindmone{\setind}} - \score_\m)^2}, \qquad \mbox{valid
  for each $\m \in \S_\setind(\Pmat)$ and $\setind > 1$.}
\label{eq:lgek}
\end{align}

Now consider an index $\m \in \S_\setind(\Pmat)$ for some $\setind <
\numsets$.  In this case, again construct an alternative pairwise
comparison matrix $\Pmat'$ under which $\m \notin \S_\setind(\Pmat')$.
Specifically, for notational convenience, we set
\begin{align}
\pmat_{ij}' =
\begin{cases}
\pmat_{\m j} - (\score_{\m} - \score_{\upindponeph}) , & i=\m, j \in
     [\numitems] \setminus \{\m\} \\ \pmat_{i \m} + (\score_{\m} -
     \score_{\upindponeph} ), & j=\m, i \in [\numitems] \setminus
     \{\m\}\\ \pmat_{ij} & \text{otherwise}.
\end{cases} \nonumber
\end{align}
In a similar manner to our earlier argument, we have $\score_i' =
\score_i + \frac{1}{\numitems-1} (\score_{\m} -
\score_{\upindponeph})$ for $i \in [\numitems] \setminus \{ \m \}$ and
$\score_{\m}' = \score_{\upindponeph}$ (relative to the scores
$\score_i$, the score of $\m$ is smaller and all others are larger by
the same factor).  Under $\Pmat'$, item $\m$ is not amongst the
$\upind{\setind}$ items with the largest scores, and therefore $\m
\notin \S_\setind(\Pmat')$.  Carrying out the same computations as
above yields:
\begin{align}
\EX[\Pmat]{\numcmp_\m} \geq \frac{\log(1/(2\delta))}{16 (\score_\m -
  \score_{\upindpone{\setind}})^2}.
\label{eq:llek}
\end{align}
Combining inequalities~\eqref{eq:lgek} and~\eqref{eq:llek} across all
items $\m$ yields the bound
\begin{align*}
\EX[\Pmat]{\numcmp} &= \sum_{i=1}^\numitems \EX{\numcmp_i} \; \geq \;
\clow \log(1/(2\delta)) \left[ \sum_{i \in \S_1} \gaplomtwo{1}{i} +
  \sum_{ \setind=2}^{\numsets - 1} \sum_{i \in \S_\setind} \max \Big
  \{ \gaplomtwo{\setind}{i}, \gapupmtwo{\setind}{i} \Big \} + \sum_{i
    \in \S_\numsets} \gapupmtwo{\numsets}{i} \right],
\end{align*}
with $\clow = 1/16$, thereby yielding the claimed result.


\subsection{Proof of Theorem~\ref{thm:necessityparametric}(a) } 

\renewcommand\loindmonemh{k}

Our goal is to prove that any algorithm $\alg$ that is uniformly
$\delta$-accurate over $\parfamily[\pmatmin]$, when applied to a given
pairwise comparison model $\Pmat \in \parfamily[\pmatmin]$, must make
at least
\begin{align*}
\EX[\Pmat]{\numcmp} & \geq \frac{\pmatmin \pdfmin^2}{2.004\pdfmax^2}
\log \left(\frac{1}{2\delta} \right) \complexityP(\vscore(\Pmat))
 \end{align*}
comparisons on average. Here $\complexityP(\vscore(\Pmat))$ is the
complexity parameter defined in equation~\eqref{def:loboSC}.

The proof is similar to that of Theorem~\ref{ThmRanking}(b), with the
primary difference being that the alternative matrix $\Pmat'$ must now
be constructed such that it lies in the parametric class.  In what
follows, we show how to modify the proof of Theorem~\ref{ThmRanking}
at appropriate positions in order to accommodate this difference.

Consider any parametric pairwise comparison matrix $\Pmat \in
\parfamily[\pmatmin]$. Then there exists a parameter vector $\vw \in
\reals^\numitems$ such that $\pmat_{ij} = \MYCDF(\parw_i - \parw_j)$.
By the assumption $\score_1 > \ldots > \score_\numitems$, this parameter
vector obeys $\parw_1> \ldots > \parw_\numitems$.  Consider an item
$\m \in \S_\setind(\Pmat), \setind > 1$, and set $\loindmonemh \defeq
\loindmone{\setind}$, for notational convenience.  We construct an
alternative matrix $\Pmat' \in \parfamily[\pmatmin]$ as follows.
Consider some scalar value $\rho$ that lies in the interval $0 < \rho
< \parw_{\loindmonemh} - \parw_{\loindmonemh-1}$.  Define a set of
alternative parameters as
\begin{align*}
\parw_i' & \defeq \begin{cases} \parw_\loindmonemh& \qquad \mbox{if
    $i=m$}, \\ \parw_\loindmonemh - \rho & \qquad \mbox{if $i =
    \loindmonemh$}, \\ \parw_i & \qquad \mbox{otherwise}.
\end{cases}
\end{align*}
Now let $\Pmat'$ be the matrix with pairwise comparison probabilities
$\pmat_{ij}' = \MYCDF(\parw_i' - \parw_j')$.  By definition, we have
$\parw_1 \geq \parw_i' \geq \parw_n$ for all $i \in [n]$, which
ensures that $\Pmat' \in \parfamily[\pmatmin]$.  Moreover, by
definition, item $\m$ is among the top $\loindmonemh$ items, so that
$\m \notin \S_\setind(\Pmat')$.  Since (by assumption) algorithm
$\alg$ is uniformly $\delta$-accurate over $\parfamily[\pmatmin]$, we
have both $\PR[\Pmat]{\event} \geq 1 - \delta$ and
$\PR[\Pmat']{\event} \leq \delta$, which ensures that
inequality~\eqref{eq:lbdApl} holds.  Here $\event$ denotes the
previously defined event~\eqref{eq:performancecriterioninnecessity}
that the algorithm $\alg$ correctly recovers the set structure.

Next consider the total number of comparisons of item $\m$ with all
others items, denoted by $\numcmp_\m$.  As in
inequality~\eqref{eq:bylbdApl}, we are guaranteed that
\begin{align}
\max_{j \in [\numitems] \setminus \{ \m \}} \kl(\pmat_{\m j},
\pmat_{\m j}') \EX[\Pmat]{\numcmp_\m}
& \geq \sum_{j \in [\numitems]\setminus\{ \m \}}
\EX[\Pmat]{\numcmp_{\m j}} \kl(\pmat_{\m j}, \pmat_{\m j}') \nonumber
\\
& \mystackrel{(i)}{=} \sum_{i=1}^\numitems \sum_{j=i+1}^\numitems
\EX[\Pmat]{\numcmp_{ij}} \kl(\pmat_{ij}, \pmat_{ij}') - \sum_{j \in
  [\numitems] \setminus \{ k, \m \}} \EX[\Pmat]{\numcmp_{jk}}
\kl(\pmat_{jk}, \pmat_{jk}')
\nonumber \\
&\mystackrel{(ii)}{\geq} \kl(\PR[\Pmat]{\event}, \PR[\Pmat']{\event})
-0.001 \kl(\PR[\Pmat]{\event}, \PR[\Pmat']{\event}) \nonumber \\
&\mystackrel{(iii)}{\geq} \kl(\PR[\Pmat]{\event}, \PR[\Pmat']{\event})
- \sum_{j \in
  [\numitems] \setminus \{ k, \m \}} \EX[\Pmat]{\numcmp_{jk}}
\kl(\pmat_{jk}, \pmat_{jk}') \nonumber \\
\label{eq:bylbdApl2}
&\geq 0.999 \log \frac{1}{2\delta} .
\end{align}
Here inequality~(i) follows from the fact that $\kl(\pmat_{ij},
\pmat_{ij}')=0$ for all $(i,j)$ with $i,j \in [\numitems]\setminus
\{k,\m\}$, by definition of $\pmat'$. Inequality~(ii) follows from
inequality~\eqref{eq:lemspecialized} (that is, from
Lemma~\ref{lem:changemeasure}). Inequality (iii) is a result of the
fact that $\lim_{\rho \to 0} \kl(\pmat_{ik},\pmat_{ik}') = \lim_{\rho
  \to 0} \kl(\MYCDF(\parw_i - \parw_k), \MYCDF(\parw_i - \parw_k+\rho)
) = 0$ for every $i \in [\numitems]\setminus \{k,\m\}$, where we have
also employed the continuous mapping theorem: Due to this relation we
can choose $\rho$ sufficiently close to $0$ to obtain the bound of
(iii).  Finally, inequality~\eqref{eq:bylbdApl2} is a consequence of
inequality~\eqref{eq:lbdApl}.

Our next step is to upper bound the KL divergence $\kl(\pmat_{\m j},
\pmat_{\m j}')$ For each $j \in [\numitems] \setminus \{k, \m \}$, we
have
\begin{align}
\kl(\pmat_{\m j}, \pmat_{\m j}') & \leq \frac{(\pmat_{\m j}' -
  \pmat_{\m j})^2}{\pmat_{\m j}' (1-\pmat_{\m j}') } \nonumber \\
& \mystackrel{(i)}{=} \frac{(\pmat_{\loindmonemh j} - \pmat_{\m
    j})^2}{\pmat_{\loindmonemh j} (1-\pmat_{\loindmonemh j}) }
\nonumber \\
& \mystackrel{(ii)}{\leq} \frac{2}{\pmatmin} (\MYCDF(\parw_\loindmonemh
- \parw_j ) - \MYCDF(\parw_\m - \parw_j ))^2 \nonumber \\
& \mystackrel{(iii)}{\leq} \frac{2}{\pmatmin} (\pdfmax
(\parw_\loindmonemh - \parw_\m ))^2 \nonumber \\
& \leq \frac{2\pdfmax^2}{\pmatmin \pdfmin^2} (\score_\loindmonemh -
\score_\m)^2.
\label{eq:userelwtau} 
\end{align}
Here step~(i) follows by definition of the parameters $\pmat_{ij}'$;
step~(ii) follows because $\pmat_{ij}$ belongs to the interval
$[\pmatmin , 1 - \pmatmin]$; and step~(iii) is a consequence of assumption~\eqref{eq:assumptionderivativeCDF}.  Finally, the last
inequality~\eqref{eq:userelwtau} follows from the relations
\begin{align}
\score_\loindmonemh - \score_\m & = \frac{1}{\numitems-1}
\left(\MYCDF(\parw_\loindmonemh - \parw_\m ) - \MYCDF(\parw_\m -
\parw_\loindmonemh) + \sum_{j \in [\numitems] \setminus
  \{\loindmonemh, \m\}} \left(\MYCDF(\parw_\loindmonemh - \parw_j ) -
\MYCDF(\parw_\m - \parw_j ) \right) \right) \nonumber \\
& \mystackrel{(i)}{\geq } \frac{1}{\numitems-1} \left(
\pdfmin(\parw_\loindmonemh - \parw_\m -( \parw_\m -
\parw_\loindmonemh)) + \sum_{j \in [\numitems] \setminus
  \{\loindmonemh, \m\}} \pdfmin (\parw_\loindmonemh - \parw_\m)
\right) \nonumber \\
& = \frac{\numitems}{\numitems - 1} \pdfmin (\parw_\loindmonemh -
\parw_\m) \geq \pdfmin (\parw_\loindmonemh - \parw_\m).
\label{eq:tauktaumrelwk}
\end{align}
Here inequality~(i) follows from
assumption~\eqref{eq:assumptionderivativeCDF}; in particular, recall
that $\parw_k > \parw_\m$, so the difference $\parw_k-\parw_\m$ above
is positive.

Similarly, we have
\begin{align}
\kl(\pmat_{\m k}, \pmat_{\m k}') \; \leq \; \frac{2}{\pmatmin} (
\pdfmax ( \rho + \parw_\loindmonemh - \parw_\m ))^2  
& \mystackrel{(i)}{\leq} \frac{2.001}{\pmatmin} ( \pdfmax (
\parw_\loindmonemh - \parw_\m ))^2 \nonumber \\
& \mystackrel{(ii)}{\leq} \frac{2.001 \pdfmax^2}{\pmatmin \pdfmin^2}
  (\score_\loindmonemh - \score_\m)^2,
\label{eq:userelwtau2} 
\end{align}
where inequality~(i) follows from choosing $\rho$ sufficiently close
to $0$, whereas inequality (ii) follows from the
relation~\eqref{eq:tauktaumrelwk}.

Given an index $\m$ in a set $\S_\setind(\Pmat)$ with $\setind > 1$,
combining inequalities~\eqref{eq:userelwtau}
and~\eqref{eq:userelwtau2} with inequality~\eqref{eq:bylbdApl2} yields
\begin{subequations}
\begin{align}
\EX[\Pmat]{\numcmp_\m} \geq \frac{\pmatmin \pdfmin^2}{2.004\pdfmax^2}
\frac{\log(1/(2\delta))}{(\score_{\loindmone{\setind}} -
  \score_\m)^2}.
\label{eq:lgekbtl}
\end{align}
Similarly, for an index $\m \in \S_\setind(\Pmat)$ with
$\setind<\numsets$, we define an alternative matrix $\Pmat'$ by
defining corresponding parameters as $\parw_\m' = \parw_\upindponeph$,
$\parw_\upindponeph'= \parw_\upindponeph + \rho$ for $\rho \in (0,
\parw_{\upind{\setind}} - \parw_{\upindponeph})$, and $\parw_i' =
\parw_i$, for all $i \notin \{\m, \upindponeph\}$.  Under the model
specified by $\Pmat'$, item $\m$ is not amongst the $\upind{\setind}$
items with the largest scores, and therefore $\m \notin
\S_\setind(\Pmat')$.  The same line of arguments as above yields
\begin{align}
\EX[\Pmat]{\numcmp_\m} \geq \frac{\pmatmin \pdfmin^2}{2.004\pdfmax^2}
\frac{\log(1/(2\delta))}{ (\score_\m - \score_{\upindpone{\setind}}
  )^2}.
\label{eq:llekbtl}
\end{align}
\end{subequations}
Combining the lower bounds~\eqref{eq:lgekbtl} and~\eqref{eq:llekbtl}
concludes the proof.


\subsection{Proof of Theorem~\ref{thm:necessityparametric}(b)}
\label{app:majorizationlemma}

Let $\vscore \in (0,1)^\numitems$ be any set of scores that is
realizable by some pairwise comparison matrix in
$\comparisonclass[\pmatmin]$. 
Theorem~\ref{thm:necessityparametric}(b) is proven by showing that 
for any continuous and strictly
increasing $\MYCDF$, there exists a pairwise comparison matrix in
$\parfamily[\pmatmin]$ with scores $\vscore$. 
As mentioned before, the proof of Theorem~\ref{thm:necessityparametric}(b) relies
on results established by Joe~\cite{joe_majorization_1988} on
majorization orderings of pairwise probability matrices.  For
convenience, we define the set of pairwise probability matrices with
scores $\vscore = (\score_1,\ldots,\score_n)$ as
\begin{align*}
\comparisonclass(\vscore) = \left\{\Pmat \in \comparisonclass[0] \,
\mid \frac{1}{\numitems-1} \sum_{j\neq i} \pmat_{ij} = \score_i,
\mbox{ for all $i$} \right\}.
\end{align*}

\paragraph{Minimality for pairwise comparison matrices:}  

Our proof requires some background on majorization and a certain
notion of minimality for pairwise comparison matrices.  We say that a
vector $\vy \in \reals^m$ is non-increasing if its entries satisfy
$y_1 \geq y_2 \geq \ldots \geq y_m$.  Given two non-increasing
vectors $\vy, \vz \in \reals^m$ such that $\sum_{i=1}^m y_i =
\sum_{i=1}^m z_i$, we say $\vy$ majorizes $\vz$, written $\vy \succ
\vz$, if
\begin{align*}
\sum_{i=1}^k y_{i} \geq \sum_{i=1}^k z_{i}, \text{ for all }
k=1,\ldots,m-1.
\end{align*}

Given pairwise comparison matrices $\Pmat, \Qmat \in
\comparisonclass(\vscore)$, we let
$\vv(\Pmat), \vv(\Qmat) \in (0,1)^{n(\numitems - 1)}$ be vectors with
entries corresponding to the off-diagonal elements of $\Pmat$ and
$\Qmat$, respectively, in non-increasing order.  We say that
\emph{$\Pmat$ majorizes $\Qmat$} if $\vv(\Pmat) \succ \vv(\Qmat)$, and
we use the shorthand $\Pmat \succ \Qmat$ to denote this relation.
Finally, a matrix $\Pmat \in \comparisonclass(\vscore)$ is
\emph{minimal} if any other $\Qmat \in \comparisonclass(\vscore)$
obeying $\Pmat \succ \Qmat$ satisfies the relation $\vv(\Qmat) =
\vv(\Pmat)$.

In order to prove Theorem~\ref{thm:necessityparametric}(a), we show
that there is a minimal $\Pmat \in \comparisonclass(\vscore) \cap
\comparisonclass[\pmatmin]$.  We first note that
Joe~\cite[Thm.~2.7]{joe_majorization_1988} observed that the argument
minimizing any \emph{Schur convex}\footnote{In our context, a function
  $f\colon (0,1)^{n\times n} \to \reals$ is Schur convex (or
  order-preserving) if for all $\Pmat, \Qmat \in
  \comparisonclass(\vscore)$ such that $\Pmat$ is majorized by
  $\Qmat$, we have $f(\Pmat) \leq f(\Qmat)$.} function over the set
$\comparisonclass(\vscore)$ is a minimal $\Pmat$.  Let us now
construct a function that is Schur convex.  In particular, we first
define a scalar function $\psi: [0,1] \rightarrow [0,\infty]$ as
\begin{align}
\psi(u) =
\begin{cases}
\frac{1}{2} \int_{1/2}^u \inv{\MYCDF}(x)dx , \quad u \in \left[\frac{1}{2},1 \right], \\
-\frac{1}{2} \int^{1/2}_u \inv{\MYCDF}(x)dx , \quad u \in \left[0,\frac{1}{2} \right) . 
\end{cases}
\end{align}
The function $\psi$ is well defined since the inverse
$\inv{\MYCDF}$ exists due to our assumption that $\MYCDF$ is strictly
increasing and continuous.  Since $\MYCDF$ is strictly increasing, so
is $\inv{\MYCDF}$.  It follows that $\psi$ is strictly convex.
From the property that all symmetric and strictly convex functions are
also strictly Schur convex, it follows that the function $\sum_{i,
  j\neq i} \psi(\pmat_{ij})$ is strictly Schur convex over
$\comparisonclass(\vscore)$.  As a result, we are guaranteed that the
argument minimizing the following convex program corresponds to a minimal matrix:
\begin{align}
\text{minimize } & \sum_{i, j>i} (\psi(\pmat_{ij}) +
\psi(1-\pmat_{ij}) ) \label{eq:primalminimial} \\ \text{subject to } &
0 \leq\pmat_{ij} \leq 1, \quad \text{ for all } i=1,\ldots,\numitems,
\quad j = i+1,\ldots,\numitems, \text{ and } \nonumber \\
 &\frac{1}{\numitems-1} \sum_{j = 1}^{i-1} (1-\pmat_{ji}) + \frac{1}{\numitems-1} \sum_{j =
  i+1}^\numitems\pmat_{ij} = \score_i, \quad
\text{ for all } i=1,\ldots,\numitems. \nonumber
\end{align}
Here the minimization is performed over the variables $\pmat_{ij}$ for
$i=1,\ldots,\numitems$ and $j=i+1,\ldots,\numitems$.

We next show that any optimal solution $\opt{\pmat}$ to the
problem~\eqref{eq:primalminimial} has entries satisfying the
interval inclusion $\opt{\pmat}_{ij} \in [\pmatmin, 1-\pmatmin]$ for all pairs $(i,j)$, and therefore $\opt{\pmat} \in \comparisonclass[\pmatmin]$, as desired. 
Indeed, suppose that there were an optimal solution $\opt{\pmat}$ that
violated this inclusion.  By assumption, there exists a matrix
$\Pmat' \in \comparisonclass(\vscore) \cap
\comparisonclass[\pmatmin]$.  Thus, if the inclusion were
violated, then there would be some index pair $(i,j)$ such that
$\opt{\pmat}_{ij} < \Pmat'_{ij}$.  This would imply that
$\opt{\Pmat}$ is strictly larger than $\Pmat'$ in the majorization
ordering.  But since the objective function~\eqref{eq:primalminimial}
is Schur convex, this contradicts the optimality of $\opt{\Pmat}$.

Since there exists a solution to the convex optimization
problem~\eqref{eq:primalminimial} that satisfies the inequality
constraints strictly (due to $\pmatmin>0$, by assumption), Slater's conditions hold, and the
Karush-Kuhn-Tucker (KKT) conditions are necessary and sufficient for
optimality (see, for instance,~\cite[Sec.~5.5]{boyd_convex_2004}).
Thus, the primal and dual optimal solutions $\opt{\pmat}_{ij}$ and
$\{\opt{\lambda}_{ij},\opt{\kappa}_{ij}, \opt{\nu}_i \}$ must satisfy
the KKT conditions
\begin{subequations}
\begin{align}
\opt{\lambda}_{ij}, \opt{\kappa}_{ij} &\geq 0, \\
\opt{\lambda}_{ij} (\opt{\pmat}_{ij} -1) = 0, \quad \opt{\kappa}_{ij}
\opt{\pmat}_{ij} &= 0, \quad \mbox{and} \quad \\
\psi'(\opt{\pmat}_{ij}) - \psi'(1- \opt{\pmat}_{ij}) +
\opt{\lambda}_{ij} - \opt{\kappa}_{ij} + \opt{\nu}_i - \opt{\nu}_j &=
0. \label{eq:KKTcondlast}
\end{align}
\end{subequations}

Since $\opt{\pmat}_{ij} \in (0,1)$ for all pairs $(i,j)$, the KKT
conditions imply that $\opt{\lambda}_{ij}=0$ and
$\opt{\kappa}_{ij}=0$.  Consequently, equation~\eqref{eq:KKTcondlast}
takes the simpler form
\begin{align}
\opt{\nu}_j - \opt{\nu}_i \; = \; \psi'(\opt{\pmat}_{ij}) -
\psi'(1-\opt{\pmat}_{ij}) & = \frac{1}{2}
\inv{\MYCDF}(\opt{\pmat}_{ij})- \frac{1}{2} \inv{\MYCDF}(1
-\opt{\pmat}_{ij}) \nonumber \\
& \mystackrel{(i)}{=} \inv{\MYCDF}(\opt{\pmat}_{ij}),
\end{align}
where step (i) follows because $\MYCDF(\timeind) = 1-
\MYCDF(-\timeind)$ for all $t\in \reals$ by assumption.  It follows
that $\opt{\pmat}_{ij} = \MYCDF(\opt{\nu}_j - \opt{\nu}_i)$ for all
pairs $(i,j)$, meaning that $\opt{\pmat}$ takes a parametric form, as
claimed.


\section{Discussion}
\label{sec:discussion}

In this paper, we considered the problem of finding a partial or
complete ranking from active pairwise comparisons.  We proved that a
simple and computationally efficient algorithm succeeds in recovering
the ranking with a sample complexity that is optimal up to logarithmic
factors.  We furthermore proved that this algorithm remains optimal
when imposing common parametric assumptions such as the popular BTL or
Thurstone models---provided the pairwise comparison probabilities are
bounded away from $0$ and $1$. This show that, perhaps surprisingly,
imposing common parametric assumptions cannot reduce the sample
complexity of ranking by more than a log-factor in the stochastic regime. 
That being said, it
should be noted that in practice, the possibility of gaining (at most)
a log factor from assuming the parametric model may be overshadowed by
the significant additional robustness afforded by our more general
model class.  For instance, see Ballinger et
al.~\cite{ballinger_decisions_1997} for some empirical evidence that
parametric models do not provide good fit in many applications, and, as
our numerical results demonstrated, algorithms relying on parametric
models can be quite sensitive to violations of those modeling
assumptions.

There are a number of open and practically relevant questions
suggested by our work.  From a theoretical perspective, it would be
interesting to provide an algorithm and corresponding guarantees for
parametric models that matches our lower bound in the regime where the
comparison probabilities are bounded away from zero and one, and at
the same time is optimal in the regime where the pairwise comparison
probabilities are very close to zero and one.  A final interesting
topic of future work is related to approximate rankings.
Specifically, in practice, one might only be interested in finding an
approximate ranking, or might only be able to find an approximate
ranking due to a limited budget of queries.


\subsection*{Acknowledgements}
The work of RH was supported by the Swiss National Science Foundation
under grant P2EZP2\_159065.  This work was supported by Office of
Naval Research MURI grant DOD-002888, Air Force Office of Scientific
Research Grant AFOSR-FA9550-14-1-001, Office of Naval Research grant
ONR-N00014, as well as National Science Foundation Grant
CIF-31712-23800.  The work of NBS was also supported in part by a
Microsoft Research PhD fellowship.


\printbibliography


\end{document}